\PassOptionsToPackage{unicode}{hyperref}
\PassOptionsToPackage{hyphens}{url}
\PassOptionsToPackage{dvipsnames,svgnames,x11names}{xcolor}
\documentclass[
]{article}
\usepackage{amsmath,amssymb}
\usepackage{iftex}
\ifPDFTeX
  \usepackage[T1]{fontenc}
  \usepackage[utf8]{inputenc}
  \usepackage{textcomp} 
\else 
  \usepackage{unicode-math} 
  \defaultfontfeatures{Scale=MatchLowercase}
  \defaultfontfeatures[\rmfamily]{Ligatures=TeX,Scale=1}
\fi
\usepackage{lmodern}
\ifPDFTeX\else
\fi
\IfFileExists{upquote.sty}{\usepackage{upquote}}{}
\IfFileExists{microtype.sty}{
  \usepackage[]{microtype}
  \UseMicrotypeSet[protrusion]{basicmath} 
}{}
\makeatletter
\@ifundefined{KOMAClassName}{
  \IfFileExists{parskip.sty}{%
    \usepackage{parskip}
  }{
    \setlength{\parindent}{0pt}
    \setlength{\parskip}{6pt plus 2pt minus 1pt}}
}{
  \KOMAoptions{parskip=half}}
\makeatother
\usepackage{xcolor}
\usepackage[margin=1in]{geometry}
\usepackage{longtable,booktabs,array}
\usepackage{calc} 
\usepackage{etoolbox}
\makeatletter
\patchcmd\longtable{\par}{\if@noskipsec\mbox{}\fi\par}{}{}
\makeatother
\IfFileExists{footnotehyper.sty}{\usepackage{footnotehyper}}{\usepackage{footnote}}
\makesavenoteenv{longtable}
\usepackage{graphicx}
\makeatletter
\def\maxwidth{\ifdim\Gin@nat@width>\linewidth\linewidth\else\Gin@nat@width\fi}
\def\maxheight{\ifdim\Gin@nat@height>\textheight\textheight\else\Gin@nat@height\fi}
\makeatother
\setkeys{Gin}{width=\maxwidth,height=\maxheight,keepaspectratio}
\makeatletter
\def\fps@figure{htbp}
\makeatother
\providecommand{\tightlist}{%
  \setlength{\itemsep}{0pt}\setlength{\parskip}{0pt}}
\newlength{\cslhangindent}
\newlength{\csllabelwidth}
\newlength{\cslentryspacingunit} 
\newenvironment{CSLReferences}[2] 
 {
  \setlength{\parindent}{0pt}
  \ifodd #1
  \let\oldpar\par
  \def\par{\hangindent=\cslhangindent\oldpar}
  \fi
  \setlength{\parskip}{#2\cslentryspacingunit}
 }%
 {}
\usepackage{calc}

\newcommand{\CSLLeftMargin}[1]{\parbox[t]{\csllabelwidth}{#1}}
\newcommand{\CSLRightInline}[1]{\parbox[t]{\linewidth - \csllabelwidth}{#1}\break}

\usepackage{xeCJK}
\setCJKsansfont{HaranoAjiGothic-Regular.otf}[BoldFont=HaranoAjiGothic-Bold.otf]
\setCJKmonofont{HaranoAjiGothic-Regular.otf}
\xeCJKDeclareCharClass{CJK}{"25CB}
\usepackage{newunicodechar}
\newunicodechar{η}{\ensuremath{\eta}}

\usepackage{caption}
\ifLuaTeX
  \usepackage{selnolig}  
\fi
\IfFileExists{bookmark.sty}{\usepackage{bookmark}}{\usepackage{hyperref}}
\IfFileExists{xurl.sty}{\usepackage{xurl}}{} 
\hypersetup{
  pdftitle={Digits in Notation Are Not Numbers: A Scoring Pitfall in PDF-to-Markdown Conversion of Japanese Technical Documents (Preliminary Evaluation)},
  pdfauthor={Mitsumasa Wada},
  colorlinks=true,
  linkcolor={Maroon},
  filecolor={Maroon},
  citecolor={Blue},
  urlcolor={Blue},
  pdfcreator={LaTeX via pandoc}}

\title{Digits in Notation Are Not Numbers: A Scoring Pitfall in
PDF-to-Markdown Conversion of Japanese Technical Documents (Preliminary
Evaluation)}
\author{Mitsumasa Wada\\Faculty of Engineering and Design, Kagawa University\\\texttt{indv0371@kagawa-u.ac.jp}}
\date{}

\begin{document}
\maketitle

\hypertarget{abstract}{%
\subsection*{Abstract}\label{abstract}}
\addcontentsline{toc}{subsection}{Abstract}

Converting PDFs to Markdown is now routine preprocessing for search and
language models, yet scoring the numbers in the output is easy to get
wrong. We evaluate MinerU, a plain-text OCR baseline (NDLOCR-Lite), and
a deterministic post-processing pipeline (anchor insertion, footnote
reordering, code-only character fixes, gated formula repair using an
independent OCR partner, and heading reconstruction) on 48 public
Japanese pages (three documents), text-layer and rasterized. Our
pre-specified number-and-unit F1 favored the OCR baseline (0.575 vs
0.397), and a disagreement check between the two engines seemed to flag
about half of the converter's number errors (54.6\%). Both were
artifacts. The scorer counted digits in the converter's notation
(image-link paths, HTML attributes, comments), which plain text lacks.
With the notation removed (a post-hoc check), the F1 scores tie (0.575
vs 0.594), and the check flags only 13\% of number errors, because most
errors are misses that both engines share. What remains: the OCR
baseline has a lower body character error rate (0.012 vs 0.053
rasterized). The pipeline raises extra-number recall (+0.145), mostly by
restoring page numbers, but lowers heading-level agreement (0.133 to
0.061) and leaves spurious formulas unchanged; its formula repair never
fires. A pre-specified precision gate for its flags (0.80, human-judged)
is met at 0.929, yet the post-processing hypothesis fails on its
improvement items. We recommend scoring numbers after removing notation
and checking critical figures against the source. By the pre-specified
criteria, the study stays a technical report.

\textbf{Keywords:} document conversion, OCR, Japanese, Markdown,
evaluation, scoring pitfall, post-processing

\hypertarget{introduction}{%
\section{Introduction}\label{introduction}}

\hypertarget{problem}{%
\subsection{Problem}\label{problem}}

Converting PDFs to Markdown is now a routine preprocessing step for
search and summarization with large language models. Pipeline tools such
as MinerU, Docling, and Marker, and direct input to vision-language
models, compete on this task
\hypertarget{cite-wang2024mineru-1}{}\protect\hyperlink{ref-wang2024mineru}{{[}1{]}},
\hypertarget{cite-auer2024docling-1}{}\protect\hyperlink{ref-auer2024docling}{{[}2{]}},
\hypertarget{cite-poznanski2025olmocr-1}{}\protect\hyperlink{ref-poznanski2025olmocr}{{[}3{]}}. The main
benchmarks that compare them, however, cover English and Chinese. The
language attribute of OmniDocBench takes three values: \texttt{en},
\texttt{simplified\_chinese}, and \texttt{en\_ch\_mixed}
\hypertarget{cite-ouyang2025omnidocbench-1}{}\protect\hyperlink{ref-ouyang2025omnidocbench}{{[}4{]}}. Japanese
appears only in evaluations on synthetic data
\hypertarget{cite-stockmark2026jasyn-1}{}\protect\hyperlink{ref-stockmark2026jasyn}{{[}5{]}},
\hypertarget{cite-sasagawa2026synthjdoc-1}{}\protect\hyperlink{ref-sasagawa2026synthjdoc}{{[}6{]}} and in an
evaluation restricted to vertical-text recognition
\hypertarget{cite-sasagawa2026vertical-1}{}\protect\hyperlink{ref-sasagawa2026vertical}{{[}7{]}}. To our knowledge,
the ability to convert real Japanese technical documents into structured
Markdown has not been measured in published work.

Converting Japanese technical documents (notifications, standards,
reports, papers) involves failures that are less conspicuous in English.
First, recognition models trained mainly on Chinese may insert
simplified Chinese characters into Japanese text. Second, look-alike
characters are confused, such as katakana ``タ'' and kanji ``夕'', or
the prolonged-sound mark ``ー'' and the kanji numeral ``一''. Third,
formula fragments appear where there is no formula. Fourth, heading
levels are lost. In addition, in technical documents a single wrong
number is costlier than a single wrong character: \texttt{235N/mm2}
becoming \texttt{235N/mm}, or \texttt{1/200} becoming \texttt{1/20},
barely changes the character error rate but breaks downstream uses such
as extracting design values.

Because a wrong number is costly, we score numbers exhaustively, token
by token. This is also where the evaluation itself can go wrong. A
layout-aware converter's Markdown contains digits that are not page
content, such as the file names of image links and HTML table
attributes, and a scorer that reads digits from the whole output counts
them as numbers. In this study the pre-specified count did exactly that.
It favored a plain-text OCR baseline on numbers and made a two-engine
disagreement check look about four times as effective as it is (54.6\%
against 13.2\% of number errors); both effects disappear when the
notation is removed (Sections 4.4, 6.1, and 6.5). We report both counts,
labeling the notation-free one as a post-hoc change.

\hypertarget{research-questions}{%
\subsection{Research questions}\label{research-questions}}

We test three hypotheses fixed in advance (\texttt{tasks/todo.md}).

\begin{itemize}
\tightlist
\item
  \textbf{H1}: A small set of deterministic post-processing rules (a
  simplified-character guard, removal of spurious formulas, heading
  correction) substantially reduces contamination by simplified and
  look-alike characters, as well as spurious formulas.
\item
  \textbf{H2}: Comparing the outputs of two engines detects most errors
  in numbers and units.
\item
  \textbf{H3}: Failures specific to Japanese (simplified-character
  contamination, confusion of look-alike characters, spurious formulas,
  flattened headings, wrong reading order in two-column layouts) also
  occur on public documents.
\end{itemize}

The decision criteria (proceed, stay a technical report, stop) were
fixed before any converter was run and recorded in commit
\texttt{42e0e50c22b2}. This report follows those criteria. Any later
change to them is stated explicitly as a post-hoc change. The criteria
were revised twice before any output was seen or scored (Section 4.7);
the last revision restated H1 for the production post-processing
pipeline evaluated in Sections 5 and 6.

\hypertarget{contributions}{%
\subsection{Contributions}\label{contributions}}

\begin{enumerate}
\def\labelenumi{\arabic{enumi}.}
\tightlist
\item
  \textbf{A scoring pitfall in number evaluation.} We score numbers with
  units exhaustively, token by token, alongside character error rate and
  table-cell agreement (Section 4). The pre-specified count favored the
  plain-text OCR baseline and made a two-engine disagreement check look
  useful, but both were artifacts of digits in the converter's notation
  (image-link paths, HTML attributes, comments). We report notation-free
  scores next to the pre-specified ones, labeled as a post-hoc change
  (Sections 4.4, 6.1, and 6.5).
\item
  \textbf{A taxonomy of failures} specific to Japanese conversion, each
  with a definition and a detection method, automatic or visual (Section
  3).
\item
  \textbf{The effect of a production post-processing pipeline and its
  ablation}: deterministic fixes where a correction is certain, and
  disagreement flagging against two partner engines for the rest, with a
  no-partner ablation (Sections 5 and 6).
\item
  \textbf{A reusable evaluation procedure}: the same pages in text-layer
  and rasterized conditions, document-level paired bootstrap, and
  transcription rules summarized in Appendix B (Section 4).
\end{enumerate}

\hypertarget{scope-and-positioning}{%
\subsection{Scope and positioning}\label{scope-and-positioning}}

This is a preliminary evaluation of about 50 pages. Its small size is a
limitation (Section 8), so we limit our conclusions to similar Japanese
technical documents. Rasterized documents are not real scans. The
replacement table of the simplified-character guard has not been
reviewed by a native Japanese speaker. Vertical writing, ruby, and old
character forms are out of scope.

Within the scope of our search, we found no earlier comparison of such
tools on Japanese technical documents that used public documents and
public code and evaluated numbers with units exhaustively. Our search
relied on web search and cannot guarantee coverage.

\hypertarget{organization}{%
\subsection{Organization}\label{organization}}

Section 2 reviews related work, Section 3 gives the failure taxonomy,
Section 4 the evaluation design, and Section 5 the post-processing.
Sections 6 to 9 present the results, discussion, limitations, and
conclusion, respectively. The appendices contain the pipeline and
partner settings, a pointer to the transcription rules, the list of
post-processing components, the documents' sources and hashes, and the
execution environment.

\hypertarget{related-work}{%
\section{Related work}\label{related-work}}

\hypertarget{benchmarks-and-metrics-for-document-parsing}{%
\subsection{Benchmarks and metrics for document
parsing}\label{benchmarks-and-metrics-for-document-parsing}}

OmniDocBench scores Markdown conversions of nine document types with
normalized edit distance for text, TEDS for tables, CDM for formulas,
and edit distance over block sequences for reading order
\hypertarget{cite-ouyang2025omnidocbench-2}{}\protect\hyperlink{ref-ouyang2025omnidocbench}{{[}4{]}},
\hypertarget{cite-zhong2020pubtabnet-1}{}\protect\hyperlink{ref-zhong2020pubtabnet}{{[}8{]}},
\hypertarget{cite-wang2025cdm-1}{}\protect\hyperlink{ref-wang2025cdm}{{[}9{]}}. olmOCR-Bench uses unit
tests that check the presence and absence of strings, reading order,
tables, and formulas
\hypertarget{cite-poznanski2025olmocr-2}{}\protect\hyperlink{ref-poznanski2025olmocr}{{[}3{]}}. READoc evaluates
multi-page conversion end to end and scores headings independently
\hypertarget{cite-li2024readoc-1}{}\protect\hyperlink{ref-li2024readoc}{{[}10{]}}. All of these target
English (and Chinese); none includes Japanese documents.

To our knowledge, FinCriticalED is the only benchmark that measures
numerical fidelity independently. It scores English financial documents
at the level of facts (numbers, dates) and shows cases where the
character error rate is low although numbers are corrupted
\hypertarget{cite-he2025fincriticaled-1}{}\protect\hyperlink{ref-he2025fincriticaled}{{[}11{]}}. It does not cover
unit tokens or full-width/half-width variants. OHRBench shows that
edit-distance-based OCR metrics do not always agree with downstream RAG
performance \hypertarget{cite-zhang2024ohrbench-1}{}\protect\hyperlink{ref-zhang2024ohrbench}{{[}12{]}}, which
motivates reporting number and structure metrics alongside
character-level ones.

CDM presupposes that formulas exist on both sides and therefore cannot
measure extra formulas (false positives)
\hypertarget{cite-wang2025cdm-2}{}\protect\hyperlink{ref-wang2025cdm}{{[}9{]}}. We count these ``spurious
formulas'' separately (Section 3).

\hypertarget{converters-and-japanese}{%
\subsection{Converters and Japanese}\label{converters-and-japanese}}

MinerU reports that rule-based pre- and post-processing contributes to
accuracy but does not isolate its contribution
\hypertarget{cite-wang2024mineru-2}{}\protect\hyperlink{ref-wang2024mineru}{{[}1{]}}. For Docling, Japanese
results depend on the OCR backend
\hypertarget{cite-auer2024docling-2}{}\protect\hyperlink{ref-auer2024docling}{{[}2{]}},
\hypertarget{cite-livathinos2025docling-1}{}\protect\hyperlink{ref-livathinos2025docling}{{[}13{]}}. YomiToku is
Japanese-specific and has no paper or published accuracy
\hypertarget{cite-yomitoku2024-1}{}\protect\hyperlink{ref-yomitoku2024}{{[}14{]}}. NDLOCR targets printed
materials from the Meiji to Showa periods and does not address Markdown
conversion of modern horizontally written technical documents
\hypertarget{cite-ndlocr_cli-1}{}\protect\hyperlink{ref-ndlocr_cli}{{[}15{]}},
\hypertarget{cite-ndlocr_lite-1}{}\protect\hyperlink{ref-ndlocr_lite}{{[}16{]}}; we therefore include
NDLOCR-Lite as a text-only OCR baseline (system B), not as a Markdown
converter.

In MinerU 3.4, Japanese was removed from the OCR language choices and
routed to a unified \texttt{ch} model
\hypertarget{cite-mineru2026software-1}{}\protect\hyperlink{ref-mineru2026software}{{[}17{]}}; PDFs with a text
layer bypass OCR. Comparing the same pages as born-digital documents
(with a text layer) and as rasterized documents therefore tests whether
simplified-character contamination arises on the OCR path. This
two-condition design is central to Section 4. Licenses also differ by
version (MinerU 3.4.5 uses the MinerU Open Source License; NDLOCR-Lite
is CC BY 4.0), so we fix versions and tabulate them in the appendix.

For Japanese document understanding, JDocQA covers 5,504 Japanese PDFs
but is evaluated by question answering and does not measure
transcription fidelity
\hypertarget{cite-onami2024jdocqa-1}{}\protect\hyperlink{ref-onami2024jdocqa}{{[}18{]}}. A study of vertical
Japanese text showed that training on synthetic data transfers poorly to
real pages \hypertarget{cite-sasagawa2026vertical-2}{}\protect\hyperlink{ref-sasagawa2026vertical}{{[}7{]}}, which
argues for evaluating on real documents. We found no public material
comparing MinerU and NDLOCR-Lite on the same Japanese technical
documents.

\hypertarget{look-alike-characters-and-simplified-characters}{%
\subsection{Look-alike characters and simplified
characters}\label{look-alike-characters-and-simplified-characters}}

Classical work on Japanese OCR post-processing models confusions with a
prior smoothed by glyph similarity
\hypertarget{cite-nagata1998japanese-1}{}\protect\hyperlink{ref-nagata1998japanese}{{[}19{]}}, and later work
corrects systematic misreadings of the same character using
document-level statistics
\hypertarget{cite-masuda2015global-1}{}\protect\hyperlink{ref-masuda2015global}{{[}20{]}}. Our guard resembles
this line of work in handling systematic substitutions with a
substitution table. We found no peer-reviewed quantification of
simplified-character contamination in Japanese output; issue reports for
PaddleOCR and MinerU describe problems with Japanese dictionaries and
accuracy but give no counts
\hypertarget{cite-paddleocr_issue11249-1}{}\protect\hyperlink{ref-paddleocr_issue11249}{{[}21{]}},
\hypertarget{cite-mineru_issue2521-1}{}\protect\hyperlink{ref-mineru_issue2521}{{[}22{]}}.

There is no settled convention for normalizing Japanese text before
computing CER. A lenient CER that tolerates orthographic variation has
been proposed \hypertarget{cite-karita2023lenient-1}{}\protect\hyperlink{ref-karita2023lenient}{{[}23{]}}; we
instead use NFKC and whitespace removal (Section 4). Because NFKC can
map some dashes to the prolonged-sound mark, we count confusions between
``ー'' and ``一'' on unnormalized strings.

\hypertarget{ocr-post-processing-and-multi-engine-agreement}{%
\subsection{OCR post-processing and multi-engine
agreement}\label{ocr-post-processing-and-multi-engine-agreement}}

OCR post-correction is conventionally divided into detection and
correction \hypertarget{cite-nguyen2021survey-1}{}\protect\hyperlink{ref-nguyen2021survey}{{[}24{]}}. In the
ICDAR 2017 competition, only about half of the methods improved on the
input text \hypertarget{cite-chiron2017icdar-1}{}\protect\hyperlink{ref-chiron2017icdar}{{[}25{]}}.
Post-correction with large language models depends strongly on the
language \hypertarget{cite-kanerva2025nofreelunches-1}{}\protect\hyperlink{ref-kanerva2025nofreelunches}{{[}26{]}}, and
a comparison of 13 PDF-to-text routes reported that LLM post-processing
gave no systematic improvement and often degraded accurate extraction
\hypertarget{cite-gomezrey2026comparative-1}{}\protect\hyperlink{ref-gomezrey2026comparative}{{[}27{]}}. These
findings are why we restrict post-processing to deterministic rules and
guarantee that it does nothing when no rule applies (Section 5).

Multi-engine combination goes back to ROVER
\hypertarget{cite-fiscus1997rover-1}{}\protect\hyperlink{ref-fiscus1997rover}{{[}28{]}} and to alignment of
several OCR outputs \hypertarget{cite-lund2009improving-1}{}\protect\hyperlink{ref-lund2009improving}{{[}29{]}};
learned combination generalizes poorly out of domain
\hypertarget{cite-lund2014generalize-1}{}\protect\hyperlink{ref-lund2014generalize}{{[}30{]}}. Voter disagreement
has also been used to route lines to humans
\hypertarget{cite-reul2018pretraining-1}{}\protect\hyperlink{ref-reul2018pretraining}{{[}31{]}}. Our disagreement
detection is closest to this last use: it does not select an output but
points to where a person should look.

\hypertarget{methodology-for-small-evaluations}{%
\subsection{Methodology for small
evaluations}\label{methodology-for-small-evaluations}}

Paired bootstrap is standard in machine translation evaluation
\hypertarget{cite-koehn2004significance-1}{}\protect\hyperlink{ref-koehn2004significance}{{[}32{]}}. The predictive
power of significance tests degrades as domains diverge
\hypertarget{cite-bergkirkpatrick2012significance-1}{}\protect\hyperlink{ref-bergkirkpatrick2012significance}{{[}33{]}}, and
underpowered comparisons are common; effect sizes and intervals should
therefore be reported together with the minimum detectable effect
\hypertarget{cite-card2020power-1}{}\protect\hyperlink{ref-card2020power}{{[}34{]}}. We report
document-level paired-bootstrap 95\% intervals (page-level for
reference) and restrict the scope of generalization. Preregistration has
been proposed for fixing hypotheses in advance in NLP
\hypertarget{cite-vanmiltenburg2021preregistering-1}{}\protect\hyperlink{ref-vanmiltenburg2021preregistering}{{[}35{]}}; we
use a lightweight variant that fixes the criteria in a specific commit.

\hypertarget{a-taxonomy-of-failures}{%
\section{A taxonomy of failures}\label{a-taxonomy-of-failures}}

\hypertarget{approach}{%
\subsection{Approach}\label{approach}}

We divide failures into character-level, structure-level, and
number-level ones. Character-level errors show up in edit distance, but
structure- and number-level errors often do not
\hypertarget{cite-he2025fincriticaled-2}{}\protect\hyperlink{ref-he2025fincriticaled}{{[}11{]}},
\hypertarget{cite-zhang2024ohrbench-2}{}\protect\hyperlink{ref-zhang2024ohrbench}{{[}12{]}}. Each failure is
detected either automatically (counted by code) or visually (classified
by a person), and the table states which. Automatically counted items
are detector outputs, not ground truth.

\hypertarget{taxonomy}{%
\subsection{Taxonomy}\label{taxonomy}}

\begin{longtable}[]{@{}
  >{\raggedright\arraybackslash}p{(\columnwidth - 6\tabcolsep) * \real{0.2500}}
  >{\raggedright\arraybackslash}p{(\columnwidth - 6\tabcolsep) * \real{0.2500}}
  >{\raggedright\arraybackslash}p{(\columnwidth - 6\tabcolsep) * \real{0.2500}}
  >{\raggedright\arraybackslash}p{(\columnwidth - 6\tabcolsep) * \real{0.2500}}@{}}
\caption{Taxonomy of failures.}\tabularnewline
\toprule\noalign{}
\begin{minipage}[b]{\linewidth}\raggedright
\#
\end{minipage} & \begin{minipage}[b]{\linewidth}\raggedright
Failure
\end{minipage} & \begin{minipage}[b]{\linewidth}\raggedright
Definition
\end{minipage} & \begin{minipage}[b]{\linewidth}\raggedright
Detection
\end{minipage} \\
\midrule\noalign{}
\endfirsthead
\toprule\noalign{}
\begin{minipage}[b]{\linewidth}\raggedright
\#
\end{minipage} & \begin{minipage}[b]{\linewidth}\raggedright
Failure
\end{minipage} & \begin{minipage}[b]{\linewidth}\raggedright
Definition
\end{minipage} & \begin{minipage}[b]{\linewidth}\raggedright
Detection
\end{minipage} \\
\midrule\noalign{}
\endhead
\bottomrule\noalign{}
\endlastfoot
F1 & Simplified-character contamination & Japanese text contains
characters used only in simplified Chinese, or kanji outside JIS X 0208
& Automatic (count per 1,000 characters); replacement table not yet
reviewed by a native speaker \\
F2 & Look-alike character confusion & Substitution among タ/夕, ー/一,
ロ/口, カ/力, ニ/二, エ/工, ヘ/へ & Visual (comparison with ground
truth); automatic counting not implemented \\
F3 & Spurious formula & \texttt{\$...\$} containing no letter or
operator, and LaTeX commands outside math, where there is no formula &
Automatic (count; expected 0 on pages without formulas) \\
F4 & Heading flattening & A heading becomes body text or its level is
lost & Automatic (recall of ground-truth headings, level agreement, rate
of false headings) \\
F5 & Wrong reading order in two columns & Left and right columns are
interleaved & Visual (counted separately, not mixed into body CER) \\
F6 & Number-and-unit error & Dropped or substituted digits, decimal
points, denominators, or units & Automatic (token-level precision,
recall, F1) \\
F7 & Table breakage & Cells split, merged, or values lost & Automatic
(precision, recall, F1 over cell strings) \\
\end{longtable}

Implementations: F1 --- \texttt{tools/simplified-guard},
\texttt{metrics.script\_contamination}; F2 --- ground truth
(\texttt{docs/transcription-rules.md}); F3 ---
\texttt{metrics.formula\_junk}; F4 --- \texttt{metrics.heading\_scores};
F5 --- \texttt{docs/gt-plan.md}; F6 --- \texttt{metrics.number\_scores};
F7 --- \texttt{metrics.table\_cell\_scores}.

F1 to F5 are the five types treated under H3; F6 is the primary metric
for H2; F7 is a secondary result.

\hypertarget{hypothesized-mechanisms}{%
\subsection{Hypothesized mechanisms}\label{hypothesized-mechanisms}}

\begin{itemize}
\tightlist
\item
  \textbf{F1}: a recognition model trained mainly on Chinese chooses
  simplified forms from a character dictionary shared with Chinese when
  reading Japanese. Because MinerU 3.4 routes Japanese to the unified
  \texttt{ch} model
  \hypertarget{cite-mineru2026software-2}{}\protect\hyperlink{ref-mineru2026software}{{[}17{]}}, we expect
  contamination in the rasterized condition and less in the text-layer
  condition. This is a hypothesis to be tested in Section 6.
\item
  \textbf{F2}: recognizers that ignore context confuse katakana and
  kanji pairs of similar shape. Prior Japanese post-processing used
  glyph similarity as a prior for such confusions
  \hypertarget{cite-nagata1998japanese-2}{}\protect\hyperlink{ref-nagata1998japanese}{{[}19{]}}.
\item
  \textbf{F3}: a formula recognizer emits non-formula regions (rules,
  symbols, digits) as formulas. Standard formula scores compare
  predicted formulas with reference ones and can mark correct output as
  wrong when it merely differs in notation
  \hypertarget{cite-wang2025cdm-3}{}\protect\hyperlink{ref-wang2025cdm}{{[}9{]}}; extra output with no
  reference counterpart is not penalized under such scores, so we count
  it directly.
\item
  \textbf{F4, F5}: errors of layout analysis and reading-order
  estimation.
\end{itemize}

\hypertarget{visual-classification}{%
\subsection{Visual classification}\label{visual-classification}}

F2 and F5 are classified visually, by one annotator. The design called
for a second pass after an interval, with agreement reported; this run
did not carry out the classification at all, so F2 and F5 are reported
as not judged (Sections 6 and 8).

\hypertarget{presenting-failure-examples}{%
\subsection{Presenting failure
examples}\label{presenting-failure-examples}}

Failure examples are short excerpts, with the document and page
identified. We do not reproduce the body text of third-party documents
at length (\texttt{docs/data-policy.md}).

\hypertarget{evaluation-design}{%
\section{Evaluation design}\label{evaluation-design}}

\hypertarget{documents}{%
\subsection{Documents}\label{documents}}

We use only documents whose license is stated and permits research use
and publication of results. Textbooks, training materials, and
confidential documents are excluded. The three selected documents total
48 pages (Table 2).

\begin{longtable}[]{@{}
  >{\raggedright\arraybackslash}p{(\columnwidth - 10\tabcolsep) * \real{0.1667}}
  >{\raggedright\arraybackslash}p{(\columnwidth - 10\tabcolsep) * \real{0.1667}}
  >{\raggedright\arraybackslash}p{(\columnwidth - 10\tabcolsep) * \real{0.1667}}
  >{\raggedright\arraybackslash}p{(\columnwidth - 10\tabcolsep) * \real{0.1667}}
  >{\raggedright\arraybackslash}p{(\columnwidth - 10\tabcolsep) * \real{0.1667}}
  >{\raggedright\arraybackslash}p{(\columnwidth - 10\tabcolsep) * \real{0.1667}}@{}}
\caption{Evaluation documents. All three have a text layer. Source URL,
retrieval date, and SHA-256 are in Appendix D.}\tabularnewline
\toprule\noalign{}
\begin{minipage}[b]{\linewidth}\raggedright
ID
\end{minipage} & \begin{minipage}[b]{\linewidth}\raggedright
Document
\end{minipage} & \begin{minipage}[b]{\linewidth}\raggedright
Type
\end{minipage} & \begin{minipage}[b]{\linewidth}\raggedright
Pages
\end{minipage} & \begin{minipage}[b]{\linewidth}\raggedright
License
\end{minipage} & \begin{minipage}[b]{\linewidth}\raggedright
Characteristics
\end{minipage} \\
\midrule\noalign{}
\endfirsthead
\toprule\noalign{}
\begin{minipage}[b]{\linewidth}\raggedright
ID
\end{minipage} & \begin{minipage}[b]{\linewidth}\raggedright
Document
\end{minipage} & \begin{minipage}[b]{\linewidth}\raggedright
Type
\end{minipage} & \begin{minipage}[b]{\linewidth}\raggedright
Pages
\end{minipage} & \begin{minipage}[b]{\linewidth}\raggedright
License
\end{minipage} & \begin{minipage}[b]{\linewidth}\raggedright
Characteristics
\end{minipage} \\
\midrule\noalign{}
\endhead
\bottomrule\noalign{}
\endlastfoot
c01 & Notification No.~1461 of the Ministry of Construction & legal text
& 4 & Copyright Act, Art. 13 (notifications of the state are not the
subject of copyright) & numbers almost all in kanji numerals;
headings \\
c06 & J-STAGE survey of academic societies (2024), slide-style report &
survey report & 35 & CC BY 4.0 & many charts and percentages \\
c08 & Damage-state evaluation of unbonded PCaPC columns (Concrete
Research and Technology, vol.~37, pp.~47-55) & engineering paper (two
columns) & 9 & CC BY 4.0 & equations (1)-(9), five tables with units \\
\end{longtable}

Source URLs, retrieval dates, and SHA-256 hashes are recorded in
\texttt{data/manifest.csv}. PDFs are not redistributed; they are
specified by URL and hash.

Some documents were excluded. The collation table of Ministry of Land,
Infrastructure, Transport and Tourism Notification No.~594 of 2007 (c02)
was found on inspection to be set vertically and was removed under our
policy of excluding vertical writing (2026-10-31), which reduced the set
from 50 to 48 pages. Out-of-copyright reports from the National Diet
Library were also excluded because of their vertical writing and old
character forms.

c06 is a slide-style survey report and not a technical document in the
narrow sense. We treat it as ``Japanese material rich in charts and
numbers''.

\hypertarget{two-conditions}{%
\subsection{Two conditions}\label{two-conditions}}

Each document is evaluated in two conditions (Figure 1).

\begin{itemize}
\tightlist
\item
  \textbf{Text layer}: the original PDF.
\item
  \textbf{Rasterized}: the original PDF converted to an image-only PDF
  at 300 dpi (\texttt{tools/prep/rasterize.py}). Document IDs get the
  suffix \texttt{i} (c01i, c06i, c08i).
\end{itemize}

Rasterized documents are not real scans: they have no stains, skew, or
shadows, so this condition measures the best case for images (a lower
bound on difficulty). Because the two conditions have identical content,
differences can be read as the contribution of the OCR path (Section 2).

Conversion is done one page at a time
(\texttt{tools/prep/split\_pages.py}) so that outputs align by page: 48
pages in two conditions give 96 PDFs to convert.

\hypertarget{systems-compared}{%
\subsection{Systems compared}\label{systems-compared}}

\begin{longtable}[]{@{}
  >{\raggedright\arraybackslash}p{(\columnwidth - 4\tabcolsep) * \real{0.3333}}
  >{\raggedright\arraybackslash}p{(\columnwidth - 4\tabcolsep) * \real{0.3333}}
  >{\raggedright\arraybackslash}p{(\columnwidth - 4\tabcolsep) * \real{0.3333}}@{}}
\caption{Systems compared.}\tabularnewline
\toprule\noalign{}
\begin{minipage}[b]{\linewidth}\raggedright
Label
\end{minipage} & \begin{minipage}[b]{\linewidth}\raggedright
System
\end{minipage} & \begin{minipage}[b]{\linewidth}\raggedright
Fixed items
\end{minipage} \\
\midrule\noalign{}
\endfirsthead
\toprule\noalign{}
\begin{minipage}[b]{\linewidth}\raggedright
Label
\end{minipage} & \begin{minipage}[b]{\linewidth}\raggedright
System
\end{minipage} & \begin{minipage}[b]{\linewidth}\raggedright
Fixed items
\end{minipage} \\
\midrule\noalign{}
\endhead
\bottomrule\noalign{}
\endlastfoot
A & MinerU (pipeline backend) & version 3.4.5, settings, run date \\
B & NDLOCR-Lite & version, run date; plain-text output (no Markdown
structure), included as an OCR baseline \\
E & Proposed: production post-processing pipeline applied to A & Section
5; rasterized condition only \\
Ablation & E without partner engines & Section 5 \\
\end{longtable}

\begin{figure}
\hypertarget{fig:design}{%
\centering
\includegraphics{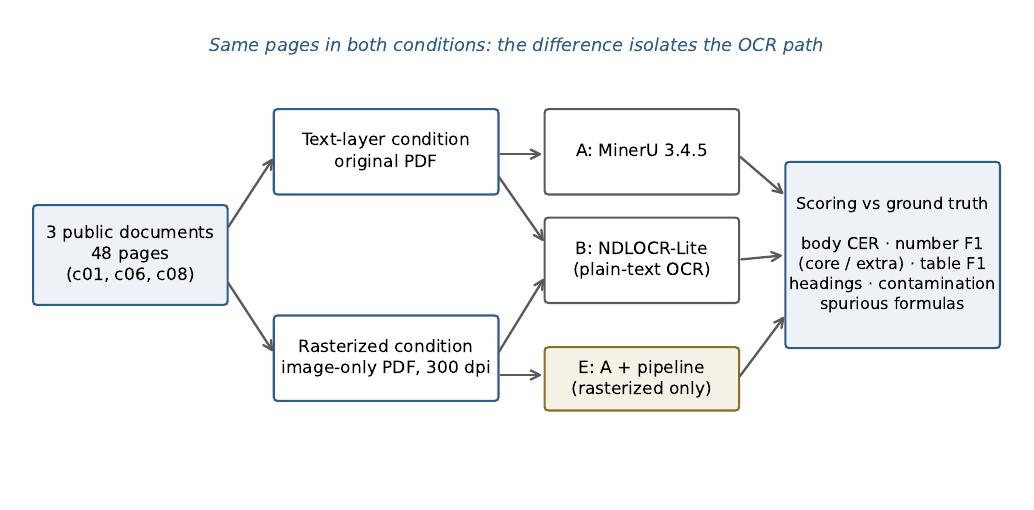}
\caption{Evaluation design. Each document is converted in two conditions
(with its text layer, and rasterized to images); the difference between
the conditions is read as the contribution of the OCR path. E is
constructed in the rasterized condition only.}\label{fig:design}
}
\end{figure}

\hypertarget{metrics}{%
\subsection{Metrics}\label{metrics}}

The number-and-unit token F1 is reported for every system as the main
descriptive result, but it is \textbf{not} the decision criterion (see
the decision rule below). Decisions use only the metrics named there;
everything else is secondary. Separating them avoids multiple-comparison
problems.

\begin{longtable}[]{@{}
  >{\raggedright\arraybackslash}p{(\columnwidth - 2\tabcolsep) * \real{0.5000}}
  >{\raggedright\arraybackslash}p{(\columnwidth - 2\tabcolsep) * \real{0.5000}}@{}}
\caption{Metrics.}\tabularnewline
\toprule\noalign{}
\begin{minipage}[b]{\linewidth}\raggedright
Metric
\end{minipage} & \begin{minipage}[b]{\linewidth}\raggedright
Definition
\end{minipage} \\
\midrule\noalign{}
\endfirsthead
\toprule\noalign{}
\begin{minipage}[b]{\linewidth}\raggedright
Metric
\end{minipage} & \begin{minipage}[b]{\linewidth}\raggedright
Definition
\end{minipage} \\
\midrule\noalign{}
\endhead
\bottomrule\noalign{}
\endlastfoot
Body CER & Edit distance between the ground-truth excerpt and the
best-matching substring of the output (semi-global alignment), divided
by the length of the excerpt; computed after NFKC normalization and
whitespace removal \\
Number-and-unit F1 & A number (with decimals, thousands separators,
fractions) and the unit immediately following it form one token;
multiset precision, recall, and F1 between ground truth (body, tables,
\texttt{extra\_numbers}) and output. Only units in a fixed list
(\texttt{metrics.UNITS}) are recognized; only Arabic numerals are
counted, and kanji numerals are evaluated through body CER \\
Number recall by part & Recall of ground-truth number tokens from the
body excerpt and tables (``core'') and from \texttt{extra\_numbers},
which holds axis ticks and equation, reference, and page numbers
(``extra''), each against the whole output multiset; reported separately
because tick labels dominate some pages (e.g.~c08 p6) \\
Table-cell F1 & Precision, recall, and F1 over cell strings; HTML and
Markdown tables in the output are both read \\
Simplified/non-JIS characters & Number of characters flagged per 1,000
characters of a page's output, averaged over pages (by the scorer's
detector for A and B; by the pipeline's own flagging for E); the
detection table has not been reviewed by a native speaker, so this is
detector output, not ground truth \\
Spurious formulas & Count as defined for F3 in Section 3 \\
Headings & Recall of ground-truth headings, level agreement, rate of
false headings \\
Processing time & Not recorded in this run; no speed comparison is made
(Section 8) \\
\end{longtable}

Digits that are part of a variable name, as subscripts or superscripts
(as in η0 or BD2), are counted as number tokens. The agreement check
against the re-transcription (Section 4.6) showed that this convention
depends on the annotator and can change token counts by about 20\% on
formula-heavy pages. We therefore made it an explicit rule of the ground
truth, fixed in Section 3 of \texttt{docs/transcription-rules.md}.

\textbf{Post-hoc variant of the number metrics (2026-11-10).} The number
counter reads digits from the whole output text. After the results were
seen, we found that MinerU's Markdown contains digits that are not page
content: file names of image links
(\texttt{!{[}{]}(images/\textless{}hash\textgreater{}.jpg)}), HTML table
attributes (\texttt{rowspan="3"}), and, in E, the page-anchor comment
(\texttt{\textless{}!-\/-\ pdf-page:\ 1\ \textbar{}\ 原典:\ 2\ -\/-\textgreater{}}).
The plain-text output of B has none. We therefore also report the number
metrics after removing HTML comments, HTML tags with their attributes,
and image-link paths and link targets (\texttt{metrics.strip\_markup}),
as \texttt{number\_tokens\_nomarkup} and
\texttt{number\_parts\_nomarkup}, together with the count of removed
number tokens (\texttt{markup\_number\_tokens}). The pre-specified
metrics are unchanged and remain the ones used in the decision criteria
(Section 4.7); the variant is reported next to them and is not used for
any decision. A further variant that joins digits spaced out inside
inline math (\texttt{0\ .\ 8\ 5}) was also computed; it changes the page
means by less than 0.001 and is not reported.

NFKC normalization and whitespace removal unify full-width and
half-width forms while keeping the distinction between the
prolonged-sound mark and the kanji numeral. Unifying dashes would hide
``ー/一'' confusion, so confusions are counted on strings before
normalization (F2). The 1-NED used in ICDAR competitions normalizes by
the longer length; we normalize by the reference length
\hypertarget{cite-shi2017rctw-1}{}\protect\hyperlink{ref-shi2017rctw}{{[}36{]}}.

The unit list was fixed by tabulating the units in the adopted
documents, and \texttt{rad} and \texttt{kN·m} were added before
evaluation (2026-10-31). The list is not changed during evaluation.

\hypertarget{statistics}{%
\subsection{Statistics}\label{statistics}}

Pages of one document are not independent, so the primary interval
resamples \textbf{documents} as units (document-level paired bootstrap,
10,000 resamples, seed 1)
\hypertarget{cite-koehn2004significance-2}{}\protect\hyperlink{ref-koehn2004significance}{{[}32{]}}; the page-level
interval is shown for reference. With three documents the intervals are
wide and discrete; we show them as they are and do not claim
significance \hypertarget{cite-card2020power-2}{}\protect\hyperlink{ref-card2020power}{{[}34{]}}. In fact, a
resample that draws the same document three times has probability 1/27
(3.7\%), which exceeds 2.5\%, so the two ends of the 95\% percentile
interval are simply the smallest and the largest per-document mean
difference. A document-level interval that excludes 0 therefore means
that all three documents moved in the same direction, and nothing more.
Per-document values are computed for the intervals; Section 6 reports
the aggregates. Text-layer and rasterized conditions are examined
separately. We limit generalization to similar Japanese technical
documents
\hypertarget{cite-bergkirkpatrick2012significance-2}{}\protect\hyperlink{ref-bergkirkpatrick2012significance}{{[}33{]}}. The
implementation is \texttt{tools/ocr-eval/compare.py}
(\texttt{-\/-cluster}) and \texttt{run\_eval.py}.

\hypertarget{ground-truth}{%
\subsection{Ground truth}\label{ground-truth}}

Ground truth exists for all 48 pages (\texttt{data/gt/}). The format and
rules are in the appendix. Coverage is as follows.

\begin{itemize}
\tightlist
\item
  Numbers and units: exhaustive over the whole page.
\item
  Body: one excerpt per page (the page's contiguous prose block; 112-520
  characters, except a contents page and a section divider in c06, which
  carry only a few characters); for two-column pages, an excerpt that
  does not cross columns.
\item
  Tables: cells of every table on the page.
\item
  Headings: every heading on the page; levels fixed per document.
\end{itemize}

\textbf{How it was made}: for all 48 pages, a language model corrected a
draft taken from the text layer against the page images
(\texttt{method:\ draft+model-check}). The pages were not written by a
person looking at the images alone. The name and version of the language
model were not recorded and cannot be identified. This procedure may
bias the text-layer condition in favor of systems that output the text
layer, so we state it as a limitation (Section 8).

\textbf{Quality check}: 10\% of the pages (five pages: c08 p6, c01 p2,
c08 p3, c06 p11, c06 p27) were re-transcribed from the page images
alone, without seeing the draft, and agreement was measured as character
CER and number-and-unit F1. The re-transcription was performed on
2026-11-07 by an LLM assistant, not a person, and not the one that made
the ground truth (method \texttt{manual}, procedure in
\texttt{docs/gt-plan.md}). For one of the five pages (c01 p2), the
transcriber had already read the adopted ground truth during a
repository audit, so blinding is incomplete for that page (recorded in
the data file's \texttt{notes} and in
\texttt{results/retranscription-notes.md}). Mean character CER was
0.0006 (maximum 0.0029; every page at or below 1\%), mean
number-and-unit F1 was 0.949, and table-cell F1 was 1.000
(\texttt{results/agreement.md}). The breakdown of the disagreements is
recorded in \texttt{results/retranscription-notes.md}. Because both
transcriptions were produced by models, this agreement is no substitute
for an independent human check (Section 8).

\hypertarget{lightweight-preregistration-and-decision-criteria}{%
\subsection{Lightweight preregistration and decision
criteria}\label{lightweight-preregistration-and-decision-criteria}}

Hypotheses H1 to H3 (Section 1.2) and the decision criteria were fixed
before any converter was run and recorded in commit
\texttt{42e0e50c22b2}. We then revised the criteria twice before
scoring. The last revision was made on 2026-11-08 and recorded in commit
\texttt{48a732f82068}; it came after conversion of system A had begun,
but before any output was seen or scored. It replaced system B with
NDLOCR-Lite, dropped the two configurations we had planned, and made E
the production post-processing pipeline of Section 5. The earlier
versions are kept unchanged in \texttt{tasks/todo.md}. What follows is
the final set of criteria.

The effect of post-processing is measured not by comparing systems but
by comparing E with A (raw MinerU) on the same pages, in the rasterized
condition, with intervals computed over documents. The comparison of A
and B is not used for the decision; it is reported descriptively.
\textbf{Proceed} if all three of the following hold:

\begin{enumerate}
\def\labelenumi{\arabic{enumi}.}
\tightlist
\item
  \textbf{H1} (we operationalize H1 of Section 1.2 as follows):

  \begin{enumerate}
  \def\labelenumii{(\alph{enumii})}
  \tightlist
  \item
    Heading-level agreement improves from A to E (the mean difference is
    positive; the interval is also reported. We made this reading
    explicit on 2026-11-10, after the results were seen; under the
    stricter reading that the interval excludes 0 and is positive, the
    decision is unchanged).
  \item
    The spurious-formula count decreases.
  \item
    We report detection coverage, the share of A's problems that E's
    review list flags (no threshold).
  \item
    The precision of E's flagging on a human-labeled sample (10 pages,
    drawn at random before any output was seen, with a recorded seed) is
    at least 80\%. Precision is computed over the simplified/non-JIS
    character flags and the spurious-formula flags together. H1 is met
    when (a), (b), and (d) all hold. If (d) is below 80\% or the sample
    does not exist, the item is ``unverified'' and we do not decide
    ``proceed''.
  \end{enumerate}
\item
  \textbf{No harm}: the upper end of the document-level 95\% interval of
  E − A is at least 0 (the interval does not lie entirely below 0), for
  number-and-unit F1 and for heading-level agreement. The number of real
  formulas damaged (visual check on c08) is reported. With only three
  documents this interval is coarse and the criterion has low
  sensitivity; the substantive safeguard is the visual count on c08.
\item
  \textbf{H3}: at least three of F1 to F5 are each confirmed on at least
  three pages as sourced examples, and their frequencies (per 1,000
  characters, share of pages) are reported.
\end{enumerate}

H2 (disagreement detection between the two engines) is not used for the
decision; it is reported with the three measures of Section 5.3
(detection rate, miss rate, false-alarm rate).

\textbf{Stay a technical report} if any item fails or is unverified;
this is the default. \textbf{Stop} if the final license check fails or
fewer than 30 public pages remain.

The human-labeled sample is used to report the precision and recall of
E's flagging. A person marks simplified-character contamination (per
character), spurious formulas, and swapped table values on MinerU output
of rasterized pages. The pages were drawn before any converter was run,
with seed 20261107 (2026-11-07): four pages from c08, four from c06, and
two from c01, all in the rasterized condition; the record is in
\texttt{results/sample-draw.md}. The labeling was carried out by the
author on 2026-11-08, and its results are reported in Section 6.

Any change after results are seen is stated with its reason and date and
labeled as a post-hoc change. No external timestamp deposit and no git
tag are used (decided 2026-11-08); the criteria are referenced by the
commits that record them in this repository.

\hypertarget{ethics-and-legal}{%
\subsection{Ethics and legal}\label{ethics-and-legal}}

All converters and the post-processing run locally; no document is sent
to an external API. No confidential documents are included. The final
check that each tool's terms allow publishing results is done before
publication. We do not reproduce third-party text beyond short sourced
excerpts.

\hypertarget{post-processing-and-disagreement-detection}{%
\section{Post-processing and disagreement
detection}\label{post-processing-and-disagreement-detection}}

\hypertarget{design-principles}{%
\subsection{Design principles}\label{design-principles}}

Configuration E is not a set of rules written for this evaluation. It is
the post-processing pipeline of the PDFtoMD production prototype,
applied unchanged to A's MinerU output (Figure 2). Its principles are:

\begin{enumerate}
\def\labelenumi{\arabic{enumi}.}
\tightlist
\item
  \textbf{Deterministic}: no model is called; the same input gives the
  same output.
\item
  \textbf{Fix only what is certain}: a reading is changed automatically
  only when a deterministic rule settles it, or when two independent
  partner engines agree on a reading that differs from the main
  engine's. Everything doubtful is flagged for a person, never guessed.
  This design avoids the over-correction reported for LLM
  post-processing
  \hypertarget{cite-kanerva2025nofreelunches-2}{}\protect\hyperlink{ref-kanerva2025nofreelunches}{{[}26{]}},
  \hypertarget{cite-gomezrey2026comparative-2}{}\protect\hyperlink{ref-gomezrey2026comparative}{{[}27{]}}.
\item
  \textbf{Everything is recorded}: every automatic change and every flag
  is written to a review list (\texttt{review.csv}) with its context, so
  the pipeline's actions can be audited item by item.
\end{enumerate}

\hypertarget{components}{%
\subsection{Components}\label{components}}

\begin{longtable}[]{@{}
  >{\raggedright\arraybackslash}p{(\columnwidth - 4\tabcolsep) * \real{0.3333}}
  >{\raggedright\arraybackslash}p{(\columnwidth - 4\tabcolsep) * \real{0.3333}}
  >{\raggedright\arraybackslash}p{(\columnwidth - 4\tabcolsep) * \real{0.3333}}@{}}
\caption{Components of the proposed pipeline.}\tabularnewline
\toprule\noalign{}
\begin{minipage}[b]{\linewidth}\raggedright
Component
\end{minipage} & \begin{minipage}[b]{\linewidth}\raggedright
Content
\end{minipage} & \begin{minipage}[b]{\linewidth}\raggedright
Failure addressed
\end{minipage} \\
\midrule\noalign{}
\endfirsthead
\toprule\noalign{}
\begin{minipage}[b]{\linewidth}\raggedright
Component
\end{minipage} & \begin{minipage}[b]{\linewidth}\raggedright
Content
\end{minipage} & \begin{minipage}[b]{\linewidth}\raggedright
Failure addressed
\end{minipage} \\
\midrule\noalign{}
\endhead
\bottomrule\noalign{}
\endlastfoot
Page anchors (P1) and footnotes (P2) & Page-number and footnote items in
MinerU's content list are placed back into the Markdown as page anchors
and footnotes, keeping pages aligned with the source & page
correspondence, footnotes \\
Heading hierarchy (P6) & Heading levels are rebuilt from the levels
recorded in the content list, not guessed from line shapes & F4 \\
Formula repair (P3) & A formula span is replaced only when the partner
engines agree on a different reading; otherwise the span is flagged, not
changed & F3, number errors \\
Character fixes (P4) & A small set of unambiguous character corrections
(look-alike characters in fixed contexts), each logged & F2 \\
Flagging & Suspected simplified or non-JIS characters, numbers that
disagree with a partner, and formula-like spans that cannot be confirmed
are written to the review list with context. Simplified characters are
\textbf{flagged, never replaced}: the replacement table has not been
reviewed by a native speaker, and the production design leaves that
decision to a person & F1, F3, numbers \\
\end{longtable}

\begin{figure}
\hypertarget{fig:pipeline}{%
\centering
\includegraphics{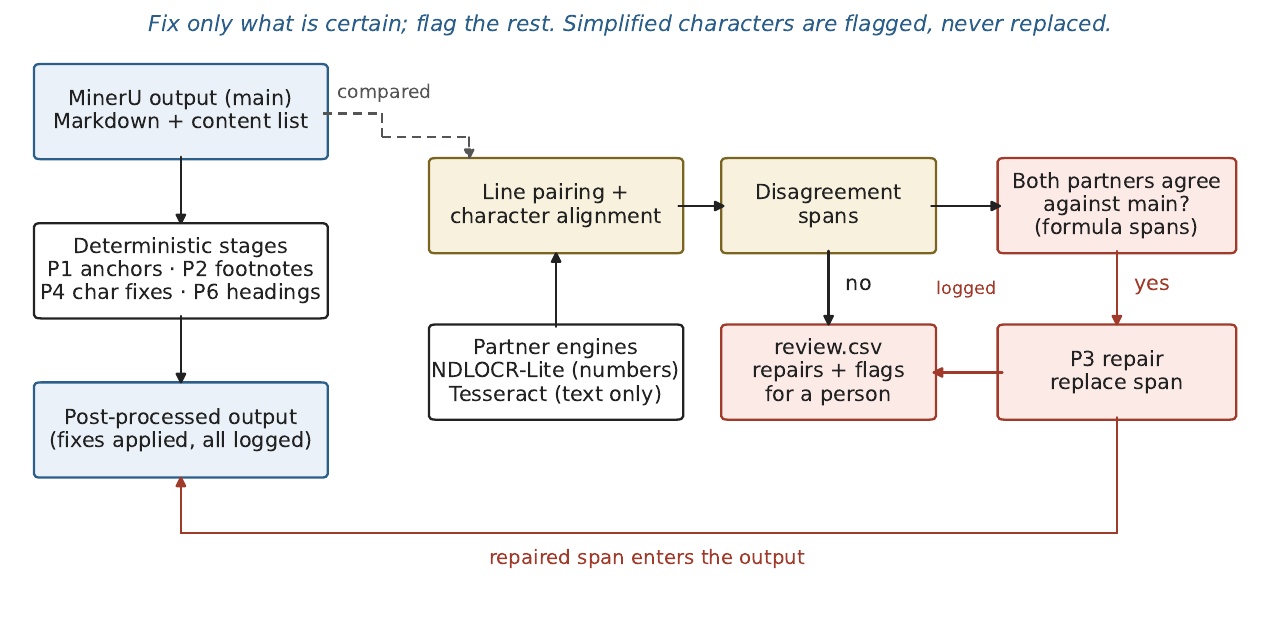}
\caption{The production post-processing pipeline (E) and its
disagreement detection. Deterministic stages transform the main output;
disagreements with the partner engines are flagged in a review list, and
a formula span is replaced only when both partners agree against the
main engine.}\label{fig:pipeline}
}
\end{figure}

Because flagging does not change the text, E does not reduce A's
simplified-character count. Its effect is therefore measured by what it
fixes (heading-level agreement, confirmed formula repairs), by how
completely it flags what it does not fix (detection coverage), and by
the precision of that flagging on a human-labeled sample (Sections 4 and
6). An earlier design of this study used three purpose-built rules (a
simplified-character guard with replacement, spurious-formula removal,
and heading correction); it was replaced by this production pipeline
before any output was seen (Section 4). That design was circular: a
removal rule was scored by a metric that shared its definition. The
circularity does not arise here. What the earlier design removed
automatically is now only flagged, and the text-changing components (the
deterministic character fixes P4 and the partner-gated formula repair
P3) are defined independently of the scorer.

\hypertarget{partner-engines-and-disagreement-detection}{%
\subsection{Partner engines and disagreement
detection}\label{partner-engines-and-disagreement-detection}}

The pipeline compares the main engine's output with two partner engines
run on the same page images: NDLOCR-Lite, used for numbers, and
Tesseract, used for running text only. Lines of the main output are
paired with the most similar partner lines, and within paired lines a
character-level alignment yields the spans where they disagree.
Differences too long to be misreadings (reading order, layout, missing
lines) are only counted as ``structural'', and table rows are compared
by numbers instead of characters. Flagging only shows where a person
should check; it does not select an output. This corresponds to ROVER's
alignment with the voting step removed and the disagreements exposed
\hypertarget{cite-fiscus1997rover-2}{}\protect\hyperlink{ref-fiscus1997rover}{{[}28{]}}, with one exception:
where both partners agree against the main engine on a formula span, P3
repairs it automatically.

System B is the NDLOCR-Lite partner's own output, scored as a converter
output, so the partner doubles as a baseline.

Three measures are reported for the disagreement detection (main engine
MinerU, partner NDLOCR-Lite; computed with
\texttt{tools/diff-check/run\_diff\_check.py} in this repository).

\begin{itemize}
\tightlist
\item
  \textbf{Detection rate}: the share of numbers that differ from ground
  truth and are flagged as disagreements.
\item
  \textbf{Miss rate}: the share of errors that raise no disagreement and
  so are not flagged; the main cause is that both engines make the same
  error. This marks where the assumption that independent errors rarely
  coincide \hypertarget{cite-wemhoener2013improved-1}{}\protect\hyperlink{ref-wemhoener2013improved}{{[}37{]}}
  fails.
\item
  \textbf{False-alarm rate}: the share of flagged disagreements where
  both outputs were in fact correct (notation differences).
\end{itemize}

\hypertarget{application-and-ablation}{%
\subsection{Application and ablation}\label{application-and-ablation}}

E applies the pipeline to A's full MinerU output (the Markdown together
with its content list) \textbf{in the rasterized condition only}. The
pipeline requires OCR-path output, but in the text-layer condition
MinerU bypasses OCR, so no content list of the required form exists.

\begin{verbatim}
python3 pipeline/run.py --mineru-dir <MinerU output for the page> --out <result folder> \
  --partner <NDLOCR-Lite text for the page> \
  --text-only-partner <Tesseract text for the page>
\end{verbatim}

The ablation (\texttt{proposed-nopartner}) is the same run with no
partners: the deterministic components (P1, P2, P4, P6) still run, but
formula repair and partner-based flagging are disabled. The pipeline
code used for the evaluation is the PDFtoMD production prototype,
version \texttt{PDFtoMD\_HighAccuracy\_proto\_v0.1.0}.

\hypertarget{results}{%
\section{Results}\label{results}}

\begin{quote}
\textbf{Executed (2026-11-08).} The converters were run on 2026-11-08 (A
and B in both conditions, E in the rasterized condition only), and
scored with \texttt{tools/ocr-eval/run\_eval.py} and
\texttt{tools/diff-check/run\_diff\_check.py}. Two execution notes
follow. (1) E was given a byte-identical copy of A's MinerU output with
the method folder renamed from \texttt{auto} to \texttt{ocr}, because
the production runner only accepts OCR-path output; A's own output was
not modified. (2) The script-contamination metric failed silently in the
first scoring run (the guard CLI did not execute), so the values below
were recomputed with the same CLI
(\texttt{tools/simplified-guard/count\_cli.js}) and cross-checked in two
environments.
\end{quote}

\hypertarget{systems-a-and-b-descriptive-not-a-decision-criterion}{%
\subsection{Systems A and B (descriptive, not a decision
criterion)}\label{systems-a-and-b-descriptive-not-a-decision-criterion}}

Overall means (page means) and document-level 95\% intervals for B − A
are shown. Section 6 reports aggregate scores. The converters' raw
outputs are not released (Section 9). Differences are computed from the
unrounded page means, so a difference may differ by 0.001 from the
difference of the rounded values shown in the tables.

\begin{longtable}[]{@{}
  >{\raggedright\arraybackslash}p{(\columnwidth - 10\tabcolsep) * \real{0.1667}}
  >{\raggedright\arraybackslash}p{(\columnwidth - 10\tabcolsep) * \real{0.1667}}
  >{\raggedright\arraybackslash}p{(\columnwidth - 10\tabcolsep) * \real{0.1667}}
  >{\raggedright\arraybackslash}p{(\columnwidth - 10\tabcolsep) * \real{0.1667}}
  >{\raggedright\arraybackslash}p{(\columnwidth - 10\tabcolsep) * \real{0.1667}}
  >{\raggedright\arraybackslash}p{(\columnwidth - 10\tabcolsep) * \real{0.1667}}@{}}
\caption{Number-and-unit F1 and recall by part (page
means).}\tabularnewline
\toprule\noalign{}
\begin{minipage}[b]{\linewidth}\raggedright
Condition
\end{minipage} & \begin{minipage}[b]{\linewidth}\raggedright
System
\end{minipage} & \begin{minipage}[b]{\linewidth}\raggedright
Body CER
\end{minipage} & \begin{minipage}[b]{\linewidth}\raggedright
Number-and-unit F1
\end{minipage} & \begin{minipage}[b]{\linewidth}\raggedright
Core recall
\end{minipage} & \begin{minipage}[b]{\linewidth}\raggedright
Extra recall
\end{minipage} \\
\midrule\noalign{}
\endfirsthead
\toprule\noalign{}
\begin{minipage}[b]{\linewidth}\raggedright
Condition
\end{minipage} & \begin{minipage}[b]{\linewidth}\raggedright
System
\end{minipage} & \begin{minipage}[b]{\linewidth}\raggedright
Body CER
\end{minipage} & \begin{minipage}[b]{\linewidth}\raggedright
Number-and-unit F1
\end{minipage} & \begin{minipage}[b]{\linewidth}\raggedright
Core recall
\end{minipage} & \begin{minipage}[b]{\linewidth}\raggedright
Extra recall
\end{minipage} \\
\midrule\noalign{}
\endhead
\bottomrule\noalign{}
\endlastfoot
Text layer & A (MinerU) & 0.047 & 0.410 & 0.980 & 0.295 \\
Text layer & B (NDLOCR-Lite) & 0.013 & 0.590 & 0.980 & 0.290 \\
Rasterized & A (MinerU) & 0.053 & 0.397 & 0.979 & 0.269 \\
Rasterized & B (NDLOCR-Lite) & 0.012 & 0.575 & 0.972 & 0.271 \\
\end{longtable}

\begin{longtable}[]{@{}lll@{}}
\caption{Differences between B and A, with document-level 95\%
intervals.}\tabularnewline
\toprule\noalign{}
Condition & B − A difference & Document-level 95\% interval \\
\midrule\noalign{}
\endfirsthead
\toprule\noalign{}
Condition & B − A difference & Document-level 95\% interval \\
\midrule\noalign{}
\endhead
\bottomrule\noalign{}
\endlastfoot
Text layer & number-and-unit F1 +0.180 & +0.125 to +0.188 \\
Text layer & body CER −0.034 & −0.042 to +0.006 \\
Rasterized & number-and-unit F1 +0.178 & +0.104 to +0.191 \\
Rasterized & body CER −0.040 & −0.055 to −0.001 \\
\end{longtable}

\begin{figure}
\hypertarget{fig:results}{%
\centering
\includegraphics{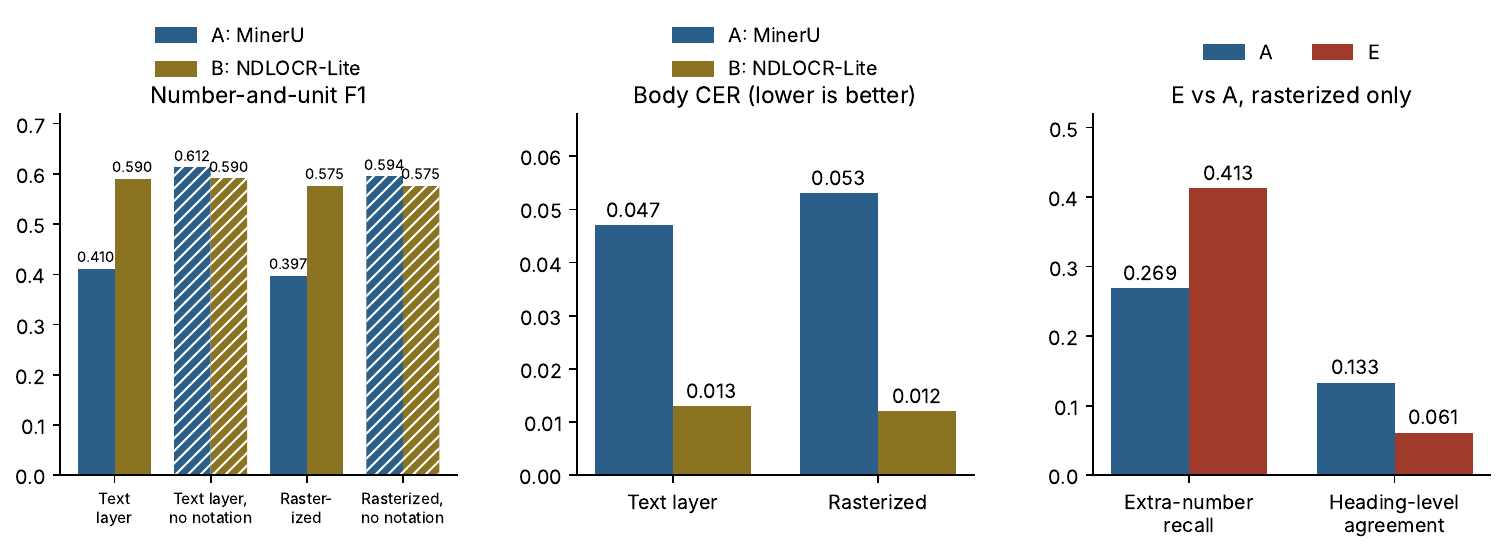}
\caption{Main results. The plain-text OCR baseline (B) reads body text
more accurately than the layout-aware converter (A) in both conditions.
Its number-and-unit F1 is higher under the pre-specified metric, but not
once notation in A's output is removed (hatched bars, post-hoc; Section
6.1). The pipeline (E, rasterized condition only) raises extra-number
recall (mostly page numbers and footnote text that it restores, Section
6.3) but lowers heading-level agreement.}\label{fig:results}
}
\end{figure}

Reading rule fixed in advance: when an interval contains 0 we write ``no
difference can be claimed''. With three documents the intervals are wide
and discrete; we report them as they are.

In the page means, B scored better than A on body text and on the
pre-specified number-and-unit F1 in both conditions (Figure 3). For that
F1 the document-level interval excludes 0 in both conditions: B is ahead
in each of the three documents (per-document table below). The next
paragraphs show that this gap comes from notation in A's output, not
from misread numbers. For body CER the interval excludes 0 only in the
rasterized condition, and only narrowly. By the reading rule, no
body-CER difference can be claimed in the text-layer condition, where A
reproduces the c01 excerpts exactly (0.000) and B does not (0.006). The
body-CER gap comes mainly from c06 (0.068 against 0.014, rasterized); on
c01 and c08 the two systems differ by 0.003 or less in the rasterized
condition.

\begin{longtable}[]{@{}
  >{\raggedright\arraybackslash}p{(\columnwidth - 14\tabcolsep) * \real{0.1250}}
  >{\raggedright\arraybackslash}p{(\columnwidth - 14\tabcolsep) * \real{0.1250}}
  >{\raggedright\arraybackslash}p{(\columnwidth - 14\tabcolsep) * \real{0.1250}}
  >{\raggedright\arraybackslash}p{(\columnwidth - 14\tabcolsep) * \real{0.1250}}
  >{\raggedright\arraybackslash}p{(\columnwidth - 14\tabcolsep) * \real{0.1250}}
  >{\raggedright\arraybackslash}p{(\columnwidth - 14\tabcolsep) * \real{0.1250}}
  >{\raggedright\arraybackslash}p{(\columnwidth - 14\tabcolsep) * \real{0.1250}}
  >{\raggedright\arraybackslash}p{(\columnwidth - 14\tabcolsep) * \real{0.1250}}@{}}
\caption{Per-document page means (c01: 4 pages; c06: 35 pages; c08: 9
pages), computed from the per-page scores.}\tabularnewline
\toprule\noalign{}
\begin{minipage}[b]{\linewidth}\raggedright
Condition
\end{minipage} & \begin{minipage}[b]{\linewidth}\raggedright
System
\end{minipage} & \begin{minipage}[b]{\linewidth}\raggedright
Body CER: c01
\end{minipage} & \begin{minipage}[b]{\linewidth}\raggedright
c06
\end{minipage} & \begin{minipage}[b]{\linewidth}\raggedright
c08
\end{minipage} & \begin{minipage}[b]{\linewidth}\raggedright
Number-and-unit F1: c01
\end{minipage} & \begin{minipage}[b]{\linewidth}\raggedright
c06
\end{minipage} & \begin{minipage}[b]{\linewidth}\raggedright
c08
\end{minipage} \\
\midrule\noalign{}
\endfirsthead
\toprule\noalign{}
\begin{minipage}[b]{\linewidth}\raggedright
Condition
\end{minipage} & \begin{minipage}[b]{\linewidth}\raggedright
System
\end{minipage} & \begin{minipage}[b]{\linewidth}\raggedright
Body CER: c01
\end{minipage} & \begin{minipage}[b]{\linewidth}\raggedright
c06
\end{minipage} & \begin{minipage}[b]{\linewidth}\raggedright
c08
\end{minipage} & \begin{minipage}[b]{\linewidth}\raggedright
Number-and-unit F1: c01
\end{minipage} & \begin{minipage}[b]{\linewidth}\raggedright
c06
\end{minipage} & \begin{minipage}[b]{\linewidth}\raggedright
c08
\end{minipage} \\
\midrule\noalign{}
\endhead
\bottomrule\noalign{}
\endlastfoot
Text layer & A & 0.000 & 0.056 & 0.032 & 0.375 & 0.376 & 0.558 \\
Text layer & B & 0.006 & 0.014 & 0.010 & 0.500 & 0.564 & 0.731 \\
Rasterized & A & 0.002 & 0.068 & 0.015 & 0.375 & 0.357 & 0.562 \\
Rasterized & B & 0.001 & 0.014 & 0.012 & 0.479 & 0.547 & 0.727 \\
Rasterized & E & 0.002 & 0.068 & 0.015 & 0.256 & 0.372 & 0.591 \\
\end{longtable}

Because c06 supplies 35 of the 48 pages, the overall page means mostly
reflect c06.

Core and extra recall are nearly equal for the two systems (core 0.97 or
above; extra 0.27 to 0.30), so the F1 difference is a precision
difference (page-mean number precision: A 0.384 and B 0.808 with the
text layer; A 0.375 and B 0.790 rasterized): A's output contains more
number tokens that match no ground-truth token. Where they come from is
shown next.

\textbf{Notation in the converter's output (post-hoc, 2026-11-10).} The
number counter reads digits from the whole output. A's Markdown contains
digits that are not page content, in file names of image links, in HTML
table attributes, and in comments; B's plain text contains none (Section
4.4). A's rasterized output has on average 29.9 such number tokens per
page (1,436 in total, on 41 of 48 pages), and E's has 32.1. Removing the
notation changes A's scores a great deal and B's not at all.

\begin{longtable}[]{@{}
  >{\raggedright\arraybackslash}p{(\columnwidth - 14\tabcolsep) * \real{0.1250}}
  >{\raggedright\arraybackslash}p{(\columnwidth - 14\tabcolsep) * \real{0.1250}}
  >{\raggedright\arraybackslash}p{(\columnwidth - 14\tabcolsep) * \real{0.1250}}
  >{\raggedright\arraybackslash}p{(\columnwidth - 14\tabcolsep) * \real{0.1250}}
  >{\raggedright\arraybackslash}p{(\columnwidth - 14\tabcolsep) * \real{0.1250}}
  >{\raggedright\arraybackslash}p{(\columnwidth - 14\tabcolsep) * \real{0.1250}}
  >{\raggedright\arraybackslash}p{(\columnwidth - 14\tabcolsep) * \real{0.1250}}
  >{\raggedright\arraybackslash}p{(\columnwidth - 14\tabcolsep) * \real{0.1250}}@{}}
\caption{Number-and-unit F1, precision, and recall (page means) before
and after removing notation. Recall is the same in the two columns for B
and changes by 0.004 or less for A and E.}\tabularnewline
\toprule\noalign{}
\begin{minipage}[b]{\linewidth}\raggedright
Condition
\end{minipage} & \begin{minipage}[b]{\linewidth}\raggedright
System
\end{minipage} & \begin{minipage}[b]{\linewidth}\raggedright
F1 (pre-specified)
\end{minipage} & \begin{minipage}[b]{\linewidth}\raggedright
F1 (notation removed)
\end{minipage} & \begin{minipage}[b]{\linewidth}\raggedright
Precision (pre-specified)
\end{minipage} & \begin{minipage}[b]{\linewidth}\raggedright
Precision (notation removed)
\end{minipage} & \begin{minipage}[b]{\linewidth}\raggedright
Recall
\end{minipage} & \begin{minipage}[b]{\linewidth}\raggedright
Number tokens in notation, per page
\end{minipage} \\
\midrule\noalign{}
\endfirsthead
\toprule\noalign{}
\begin{minipage}[b]{\linewidth}\raggedright
Condition
\end{minipage} & \begin{minipage}[b]{\linewidth}\raggedright
System
\end{minipage} & \begin{minipage}[b]{\linewidth}\raggedright
F1 (pre-specified)
\end{minipage} & \begin{minipage}[b]{\linewidth}\raggedright
F1 (notation removed)
\end{minipage} & \begin{minipage}[b]{\linewidth}\raggedright
Precision (pre-specified)
\end{minipage} & \begin{minipage}[b]{\linewidth}\raggedright
Precision (notation removed)
\end{minipage} & \begin{minipage}[b]{\linewidth}\raggedright
Recall
\end{minipage} & \begin{minipage}[b]{\linewidth}\raggedright
Number tokens in notation, per page
\end{minipage} \\
\midrule\noalign{}
\endhead
\bottomrule\noalign{}
\endlastfoot
Text layer & A & 0.410 & 0.612 & 0.384 & 0.925 & 0.493 & 29.6 \\
Text layer & B & 0.590 & 0.590 & 0.808 & 0.808 & 0.507 & 0 \\
Rasterized & A & 0.397 & 0.594 & 0.375 & 0.900 & 0.481 & 29.9 \\
Rasterized & B & 0.575 & 0.575 & 0.790 & 0.790 & 0.491 & 0 \\
Rasterized & E & 0.403 & 0.657 & 0.341 & 0.895 & 0.563 & 32.1 \\
\end{longtable}

\begin{longtable}[]{@{}
  >{\raggedright\arraybackslash}p{(\columnwidth - 8\tabcolsep) * \real{0.2000}}
  >{\raggedright\arraybackslash}p{(\columnwidth - 8\tabcolsep) * \real{0.2000}}
  >{\raggedright\arraybackslash}p{(\columnwidth - 8\tabcolsep) * \real{0.2000}}
  >{\raggedright\arraybackslash}p{(\columnwidth - 8\tabcolsep) * \real{0.2000}}
  >{\raggedright\arraybackslash}p{(\columnwidth - 8\tabcolsep) * \real{0.2000}}@{}}
\caption{Differences with and without notation.}\tabularnewline
\toprule\noalign{}
\begin{minipage}[b]{\linewidth}\raggedright
Condition
\end{minipage} & \begin{minipage}[b]{\linewidth}\raggedright
Difference
\end{minipage} & \begin{minipage}[b]{\linewidth}\raggedright
Metric
\end{minipage} & \begin{minipage}[b]{\linewidth}\raggedright
Pre-specified: difference (document-level 95\% interval)
\end{minipage} & \begin{minipage}[b]{\linewidth}\raggedright
Notation removed: difference (document-level 95\% interval)
\end{minipage} \\
\midrule\noalign{}
\endfirsthead
\toprule\noalign{}
\begin{minipage}[b]{\linewidth}\raggedright
Condition
\end{minipage} & \begin{minipage}[b]{\linewidth}\raggedright
Difference
\end{minipage} & \begin{minipage}[b]{\linewidth}\raggedright
Metric
\end{minipage} & \begin{minipage}[b]{\linewidth}\raggedright
Pre-specified: difference (document-level 95\% interval)
\end{minipage} & \begin{minipage}[b]{\linewidth}\raggedright
Notation removed: difference (document-level 95\% interval)
\end{minipage} \\
\midrule\noalign{}
\endhead
\bottomrule\noalign{}
\endlastfoot
Text layer & B − A & F1 & +0.180 (+0.125 to +0.188) & −0.022 (−0.030 to
+0.002) \\
Text layer & B − A & Precision & +0.424 (+0.167 to +0.462) & −0.117
(−0.154 to +0.000) \\
Rasterized & B − A & F1 & +0.178 (+0.104 to +0.191) & −0.019 (−0.025 to
+0.005) \\
Rasterized & B − A & Precision & +0.416 (+0.146 to +0.457) & −0.110
(−0.140 to −0.021) \\
Rasterized & E − A & F1 & +0.006 (−0.119 to +0.028) & +0.062 (−0.085 to
+0.082) \\
Rasterized & E − A & Extra recall & +0.145 (+0.000 to +0.173) & +0.146
(+0.000 to +0.174) \\
\end{longtable}

Without the notation, B no longer scores better than A on numbers. The
F1 difference is −0.019 rasterized (document-level interval −0.025 to
+0.005) and −0.022 with the text layer (−0.030 to +0.002); by the
reading rule, no difference can be claimed in either condition. Recall
is equal (B − A +0.010 and +0.013). Precision is higher for A
(rasterized: A 0.900, B 0.790; document-level interval of B − A −0.140
to −0.021). The advantage in number-and-unit F1 seen under the
pre-specified metric is therefore an artifact of counting digits in A's
notation as errors. What remains of B's advantage is body CER (the
interval excludes 0 only in the rasterized condition, and the gap comes
mainly from c06). Figure 3 shows the notation-free F1 as hatched bars
next to the pre-specified one.

\hypertarget{secondary-results}{%
\subsection{Secondary results}\label{secondary-results}}

\begin{longtable}[]{@{}
  >{\raggedright\arraybackslash}p{(\columnwidth - 12\tabcolsep) * \real{0.1429}}
  >{\raggedright\arraybackslash}p{(\columnwidth - 12\tabcolsep) * \real{0.1429}}
  >{\raggedright\arraybackslash}p{(\columnwidth - 12\tabcolsep) * \real{0.1429}}
  >{\raggedright\arraybackslash}p{(\columnwidth - 12\tabcolsep) * \real{0.1429}}
  >{\raggedright\arraybackslash}p{(\columnwidth - 12\tabcolsep) * \real{0.1429}}
  >{\raggedright\arraybackslash}p{(\columnwidth - 12\tabcolsep) * \real{0.1429}}
  >{\raggedright\arraybackslash}p{(\columnwidth - 12\tabcolsep) * \real{0.1429}}@{}}
\caption{Secondary metrics by system and condition.}\tabularnewline
\toprule\noalign{}
\begin{minipage}[b]{\linewidth}\raggedright
Condition
\end{minipage} & \begin{minipage}[b]{\linewidth}\raggedright
System
\end{minipage} & \begin{minipage}[b]{\linewidth}\raggedright
Table-cell F1
\end{minipage} & \begin{minipage}[b]{\linewidth}\raggedright
Heading-level agreement
\end{minipage} & \begin{minipage}[b]{\linewidth}\raggedright
False-heading rate
\end{minipage} & \begin{minipage}[b]{\linewidth}\raggedright
Script contamination (per 1,000 chars)
\end{minipage} & \begin{minipage}[b]{\linewidth}\raggedright
Spurious formulas (count)
\end{minipage} \\
\midrule\noalign{}
\endfirsthead
\toprule\noalign{}
\begin{minipage}[b]{\linewidth}\raggedright
Condition
\end{minipage} & \begin{minipage}[b]{\linewidth}\raggedright
System
\end{minipage} & \begin{minipage}[b]{\linewidth}\raggedright
Table-cell F1
\end{minipage} & \begin{minipage}[b]{\linewidth}\raggedright
Heading-level agreement
\end{minipage} & \begin{minipage}[b]{\linewidth}\raggedright
False-heading rate
\end{minipage} & \begin{minipage}[b]{\linewidth}\raggedright
Script contamination (per 1,000 chars)
\end{minipage} & \begin{minipage}[b]{\linewidth}\raggedright
Spurious formulas (count)
\end{minipage} \\
\midrule\noalign{}
\endhead
\bottomrule\noalign{}
\endlastfoot
Text layer & A & 0.582 & 0.133 & 0.034 & 0.020 (4 chars, 2 pages) & 1 \\
Text layer & B & 0.000 & 0.000 & 0.000 & 0.000 (0 chars) & 0 \\
Rasterized & A & 0.623 & 0.133 & 0.080 & 0.352 (16 chars, 9 pages) &
3 \\
Rasterized & B & 0.000 & 0.000 & 0.000 & 0.000 (0 chars) & 0 \\
Rasterized & E & 0.625 & 0.061 & 0.080 & 0.278 (9 chars, 6 pages) & 3 \\
Rasterized & E (no partner) & 0.625 & 0.061 & 0.080 & 0.278 (9 chars, 6
pages) & 3 \\
\end{longtable}

B's zeros for tables and headings are structural: it returns plain text
(Limitation 8). E and the no-partner ablation produced byte-identical
output files, so every metric coincides (Section 6.3). Processing time
was not recorded in this run. Script-contamination counts are detector
output, not ground truth, because the replacement table is unreviewed.
The rate is the page mean of per-page rates (Section 4.4), so it is not
proportional to the character totals in parentheses: a flagged character
on a short page weighs more than one on a long page. A's and B's counts
come from the scorer's detector and E's from the pipeline's own flagging
(Section 4.4), so E's lower count was not measured with the same
instrument as A's and does not indicate an improvement.

\hypertarget{h1-post-processing-e-against-a-rasterized-condition}{%
\subsection{H1: post-processing, E against A (rasterized
condition)}\label{h1-post-processing-e-against-a-rasterized-condition}}

\begin{longtable}[]{@{}
  >{\raggedright\arraybackslash}p{(\columnwidth - 4\tabcolsep) * \real{0.3333}}
  >{\raggedright\arraybackslash}p{(\columnwidth - 4\tabcolsep) * \real{0.3333}}
  >{\raggedright\arraybackslash}p{(\columnwidth - 4\tabcolsep) * \real{0.3333}}@{}}
\caption{H1 under the decision criteria. If the sample is missing or
precision is below 80\%, H1 is ``unverified'' and the decision is ``stay
a technical report''.}\tabularnewline
\toprule\noalign{}
\begin{minipage}[b]{\linewidth}\raggedright
Item
\end{minipage} & \begin{minipage}[b]{\linewidth}\raggedright
Criterion
\end{minipage} & \begin{minipage}[b]{\linewidth}\raggedright
Result
\end{minipage} \\
\midrule\noalign{}
\endfirsthead
\toprule\noalign{}
\begin{minipage}[b]{\linewidth}\raggedright
Item
\end{minipage} & \begin{minipage}[b]{\linewidth}\raggedright
Criterion
\end{minipage} & \begin{minipage}[b]{\linewidth}\raggedright
Result
\end{minipage} \\
\midrule\noalign{}
\endhead
\bottomrule\noalign{}
\endlastfoot
Heading-level agreement, E vs A & improves (interval reported) & 0.133
to 0.061 (difference −0.072; page-level interval −0.152 to −0.011;
document-level −0.452 to +0.000). \textbf{Did not improve} \\
Spurious formulas, E vs A & decreases & 3 to 3 (2 pages). \textbf{Did
not decrease} \\
Detection coverage: share of A's problems (simplified/non-JIS
characters, spurious formulas, number errors) flagged in E's review list
& reported (no fixed threshold) & see below \\
Flagging precision on the human-labeled sample & at least 80\% &
\textbf{0.929} (13/14; simplified 1.000, formula junk 0.889).
\textbf{Met} \\
\end{longtable}

Headings. The fall in heading-level agreement is confined to c08: four
of its seven pages with headings (p2, p4, p5, p6) dropped to 0, which
took c08 from 0.667 to 0.214. c01 (0.100) and c06 (0.029) were
unchanged.

Detection coverage. E's review list held 416 rows over 48 pages: 134
number-disagreement flags (86 misread candidates, 37 partner-only, 11
main-only), 7 simplified/non-JIS character rows, 12 spurious-formula
rows, and others. Against the ground truth, A made 636 number errors in
the pre-specified count and 333 with the notation removed. Review rows
aggregate by line, so row-level coverage, capped per page, is 87/636 =
13.7\% in the pre-specified count (not recomputed). The underlying
disagreement signal detects 13.2\% of the errors at token level once the
notation is removed (Section 6.5; 54.6\% when digits inside notation are
counted). The two figures are now of the same order; the earlier gap
between them was due to the notation. For characters, E flagged 7 rows
against the 16 suspicious characters the detector counts in A's output;
for formulas, E raised 15 formula-related rows (12 spurious-formula, 3
formula-line-count) against A's 3 spurious formulas.

Outside the criterion items, E's largest change was in extra-number
recall, which rose from 0.269 to 0.413 (difference +0.145; page-level
interval +0.093 to +0.210; document-level +0.000 to +0.173, so by the
reading rule no difference can be claimed at document level). Body CER
fell slightly (−0.0002), core recall was unchanged, and table-cell F1
was nearly unchanged (0.623 to 0.625).

The per-page scores show what this gain consists of. E matched 58 more
extra tokens than A: 33 in c06, 25 in c08, and none in c01. In c06 the
gain is exactly one token on 31 pages and two tokens on one more page.
On the two c06 pages whose only extra token is the slide's page number
(p3 and p4), extra recall went from 0 to 1. In c08, 15 of the 25 tokens
are on the title page, whose extra tokens are mostly footnote marks,
affiliation footnotes, and bibliographic data. The gain is therefore
consistent with page numbers restored by the page anchors (P1) and
footnote text restored by P2, and not with recovery of figure values.
Because recall is averaged per page, one token on a page with few extra
tokens moves the mean a great deal: c06 accounts for +0.131 of the
+0.145. We confirmed this reading directly on one page, c06 p3, whose
only extra token is the printed page number ``2''. A's output has no
page number there. E's output begins with the anchor line
\texttt{\textless{}!-\/-\ pdf-page:\ 1\ \textbar{}\ 原典:\ 2\ -\/-\textgreater{}}
followed by \texttt{**【2】**}, so the printed page number is restored
and matches the ground truth. The same anchor comment, however, adds two
more digit tokens (``1'' and ``2''), and scoring E's output on this page
gives number precision 0.80 against A's 1.00. With the comment removed
(the notation-free variant of Section 4.4), precision is 1.00 and F1 is
1.00 against A's 0.93. The notation-free scores show that the same
happens on every page: E's output has on average 2.2 more number tokens
inside notation than A's (2.1 on c06, 2.0 on c08, 4.0 on c01), which
fits an anchor comment on each page. Without the notation, E's
number-and-unit F1 is 0.657 against A's 0.594 (E − A +0.062; page-level
interval +0.026 to +0.113; document-level −0.085 to +0.082, with −0.085
on c01, +0.082 on c06, and +0.051 on c08), and extra-number recall rises
by +0.146, as before (+0.145). The gain in extra recall therefore comes
from page content, which fits the printed page number restored on c06
p3, and not from the comments. We did not re-inspect the outputs of the
other pages. On c01, number-and-unit F1 fell from 0.375 to 0.256 with
the notation counted and from 0.500 to 0.415 without it (number
precision 0.500 to 0.356). We traced this on the two c01 pages that
contain numbers (p2 and p3). E's output there begins with a page marker,
\texttt{**【c01-p02-pdf.1】**}, whose label is the name of the one-page
source file, because these pages show no printed page number. Its digits
(01, 02, 1) count as number tokens that the ground truth does not
contain; without that line, number-and-unit F1 is 1.0 on both pages,
equal to A's. The drop on c01 is therefore an effect of the pipeline's
fallback page label on our file naming, not of any correction the
pipeline makes.

The partner made no difference on this corpus: partner-gated replacement
(P3) never fired in 48 pages, so the partner's contribution appeared
only in flagging, and E's gains above come from the partner-free stages
(P1, P2, P4, P6).

The sample was labeled on 2026-11-08
(\texttt{results/sample-scores.md}). The precision gate passed, but the
heading and spurious-formula items failed their criteria, so the
improvement claims under H1 were not supported (Section 7). The
remaining flag types were also judged descriptively on the same sample
(\texttt{data/gt-sample-extra/}): all 6 deterministic character fixes
(char-fix) were correct; 7 of 23 misread candidates were real; and 0 of
24 number-only flags were real problems; most were figure-axis ticks and
notation variants (e.g.~○ vs 〇), not body-text number errors.
Line-level flags are therefore trustworthy for simplified characters and
formula junk, but number flags sent straight to a human review queue
would cost more checking effort than they return; the 13.7\% coverage
figure above should be read together with this precision. This also
reconciles two figures that can look contradictory: the token-level
false-alarm rate of the disagreement check is 0.000 (Section 6.5), yet
none of the 24 judged number flags was a real problem. The first counts
a disagreement as a false alarm only when both engines render the same
value correctly; the judges marked figure-axis ticks and notation
variants as not real problems even where one engine had genuinely added
or missed a token. The disagreement signal reliably shows that the two
engines differ; whether the differing token matters for the document's
content is a separate judgment.

\hypertarget{no-harm-criterion-2}{%
\subsection{No harm (criterion 2)}\label{no-harm-criterion-2}}

The document-level 95\% interval of E − A is −0.119 to +0.028 for
number-and-unit F1 and −0.452 to +0.000 for heading-level agreement:
neither is entirely below 0, so the no-harm criterion is met at document
level, although heading agreement does fall at page level (Section 6.3).
With three documents, the ends of each interval are the smallest and the
largest per-document difference (Section 4.5): number-and-unit F1
changed by −0.119 on c01 (caused by the page marker, Section 6.3),
+0.015 on c06, and +0.028 on c08, and heading-level agreement by 0.000
on c01 and c06 and −0.452 on c08. The criterion is thus met as soon as
one document does not get worse, which shows how weak it is. For
formulas, the criterion names a visual check on c08. The author carried
out this check and found no damaged formula (0). The records agree with
it. Formula replacement (P3) never fired, so no formula was replaced.
The only text-changing step that ran, the character fixes (P4), was
judged on the four sampled c08 pages: all six fixes were correct, and
none lay inside a formula (\texttt{data/gt-sample-extra/}). Body CER is
identical for A and E on each of the nine c08 pages.

\hypertarget{h2-disagreement-detection}{%
\subsection{H2: disagreement
detection}\label{h2-disagreement-detection}}

The disagreement check against NDLOCR-Lite, scored with the notation
removed from both texts (\texttt{-\/-strip-markup}), detected 13.2\% of
A's number errors in the rasterized condition (44 of 333), with a miss
rate of 86.8\%; in the text-layer condition the figures are 15.0\% (49
of 327) and 85.0\%. The false-alarm rate (notation-only pairs) was 0.000
in both conditions (100 flags in the rasterized condition, 99 in the
text-layer condition). No threshold was set for H2 in advance; against
its wording, detecting ``the majority,'' the measured rate falls far
short. The detections concentrate on a few pages (c08 p1, p3, and p5 and
c06 p5 account for 36 of the 44 in the rasterized condition), and the
misses concentrate on c06 (197 of 289) and on the number-dense c08 p6
(62).

The pre-specified run, which counts digits inside notation, gave 54.6\%
(347 of 636) and 53.9\% (325 of 603), with 403 and 375 flags. The whole
difference is notation. In the rasterized condition, the 303 error
tokens that the removal takes out of the count (636 − 333) were all
counted as detected (347 − 44), because B's plain text never contains
them; the text-layer condition is the same (276 each). The pre-specified
results are kept in \texttt{results/diff-check-image.md} and
\texttt{results/diff-check-textlayer.md}, and the notation-free ones in
the two files ending in \texttt{-nomarkup}.

\hypertarget{h3-reproduction-of-failures}{%
\subsection{H3: reproduction of
failures}\label{h3-reproduction-of-failures}}

F1 (simplified-character contamination) reproduced in the rasterized
condition: 16 characters on 9 of A's pages (0.352 per 1,000 characters),
meeting the three-page criterion; in the text-layer condition it was
weak (4 characters on 2 pages), and B showed none in either condition.
F4 (flattened headings) reproduced: on the 44 pages with headings, A's
level agreement stayed at 0.133 in both conditions, and false headings
also appeared (rates 0.034 and 0.080). F3 (spurious formulas) occurred
but did not reach the page criterion (3 items on 2 pages rasterized; 1
item on 1 page with the text layer). F2 (look-alike character confusion)
and F5 (two-column reading order) were not classified in this run, so no
reproduction judgment is made for them. Two types (F1 and F4) reach the
three-page criterion; criterion 3 requires three types, so it is not
met.

Sourced examples are taken from the human-labeled sample
(\texttt{data/gt-sample/}; rasterized condition, A's text as passed
through E). F1: in c06 p3, 調査対象 (``survey target'') appears with 査
replaced by the Chinese form U+67E5; in c08 p4, 非線形弾性型
(``nonlinear elastic type'') appears with 弾 replaced by the simplified
form U+5F39; in c01 p4, 掲げる (``to list'') appears with 掲 replaced by
U+63ED. The labeler confirmed contamination on five of the ten sampled
pages (c01 p4, c06 p3, c08 p2, p4, p5). F3: in c08 p4, the particle ので
in the running text ``\ldots と判定されているので実質的に\ldots{}'' was
emitted as a formula,
\texttt{\$\textbackslash{}mathcal\{O\}\textbackslash{}overline\{\textbackslash{}mathrm\{\textasciitilde{}C\textasciitilde{}\}\}\$}.
For F4 we give no excerpt, because the converters' raw outputs are not
released; its evidence is the heading scores above.

\hypertarget{effect-of-rasterization}{%
\subsection{Effect of rasterization}\label{effect-of-rasterization}}

For A, rasterization raised body CER from 0.047 to 0.053 and script
contamination from 0.020 per 1,000 characters (4 characters, 2 pages) to
0.352 (16 characters, 9 pages): the contamination arises mainly on the
rasterized path, consistent with the hypothesis that it enters through
OCR (Section 3). For B, contamination was 0 in both conditions, which is
also consistent with a Japanese-specific recognizer lacking that route.
This is an observation on three documents, not proof of causation.

\hypertarget{discussion}{%
\section{Discussion}\label{discussion}}

\hypertarget{decisions-on-h1-to-h3}{%
\subsection{Decisions on H1 to H3}\label{decisions-on-h1-to-h3}}

We apply the criteria of Section 4.7 mechanically. \textbf{H1 was not
supported.} Flagging precision on the human-labeled sample was 0.929,
which passed the gate (criterion: at least 80\%), but the two
improvements the pipeline targeted failed their criteria. Heading-level
agreement did not improve (0.133 to 0.061), and spurious formulas did
not decrease (3, unchanged). The pipeline was not without effect,
however: extra-number recall rose by +0.145 and body CER did not worsen.
Its measurable benefit appeared not in the targeted headings and
formulas but in extra-number recall, and Section 6.3 shows that this
gain is mostly page numbers and footnote text that the pipeline puts
back (P1, P2). Such tokens matter less to a reader than the values in
figures and tables.

\textbf{H2 is assessed descriptively, because no threshold was set in
advance.} With the notation removed, the disagreement check detected
13.2\% of number errors (rasterized) and 15.0\% (text layer), with no
false alarms at token level. The pre-specified count, which includes
digits inside notation, gave 54.6\% and 53.9\% (Section 6.5). Neither is
``most'', and the notation-free rate falls far short. The miss rate of
about 85\% is the main result: most number errors are ones that both
engines make, and the check cannot flag them. Core recall is 0.97 or
above for both engines (Section 6.1), so the shared misses are mostly
extra numbers such as figure ticks; this is an inference from the recall
figures, not a count. The check should be treated as a weak detector,
not as a guarantee.

\textbf{H3 is partly supported.} F1 (simplified-character contamination)
and F4 (flattened headings) reproduced on public documents. F3 (spurious
formulas) occurred, but below the preset page-count criterion. F2 and F5
were not classified, so no judgment is made on them. Two types reach the
page-count criterion where criterion 3 requires three, so criterion 3 is
not met.

\textbf{Overall decision.} Criterion 1 (H1) is not met, criterion 2 (no
harm) is met at document level, and criterion 3 (H3) is not met. Under
Section 4.7 the outcome is the default: stay a technical report.

\hypertarget{where-post-processing-did-not-help-or-did-harm}{%
\subsection{Where post-processing did not help, or did
harm}\label{where-post-processing-did-not-help-or-did-harm}}

The fall in heading-level agreement is a harm attributable to the
post-processing. The page-level interval excludes 0 (−0.152 to −0.011).
The document-level interval is wide (−0.452 to +0.000), and the
per-document values show why: the whole fall comes from c08, the
two-column engineering paper, where agreement dropped from 0.667 to
0.214 (four of its seven pages with headings fell to 0). c01 and c06
were unchanged. Hierarchy reconstruction thus backfired on the one
document whose heading levels the converter had mostly got right, and it
did not help on c06, where A's agreement was already close to zero
(0.029). Where the converter's own levels are good, reconstruction
should be weakened or the converter's headings should take precedence.

Spurious formulas did not decrease partly because A produced few to
begin with (3 in 48 pages). Evaluating formula cleanup requires a corpus
richer in formulas.

That the partner contributed only flags is also a design finding. The
condition that triggers replacement (agreement by the partner) was never
met on this corpus. Loosening it would produce more replacements, at the
price of more wrong ones. Within this study, the policy of ``fixing only
what is certain and flagging the rest'' left body text unharmed
(body-CER difference −0.0002, core recall unchanged), but it also left
fixable errors uncorrected.

\hypertarget{differences-by-document-and-condition}{%
\subsection{Differences by document and
condition}\label{differences-by-document-and-condition}}

On the pre-specified number-and-unit F1, B (NDLOCR-Lite) scored better
than A in all three documents and in both conditions, but this advantage
disappears once notation is removed from A's output (Section 6.1). On
body text its advantage is large only on c06, and it is absent on c01 in
the text-layer condition (Section 6.1). The Japanese-specific recognizer
reads body text at least as well as the converter but returns no
structure (headings, tables). Converter and OCR engine are thus
complementary rather than interchangeable, and B's two roles here, as
baseline and as detection partner, fit that division of labor.

The effect of rasterization is as reported in Section 6.7, and it
strengthens the view that simplified-character contamination is tied to
the OCR path. The number-and-unit metrics effectively measure c06 and
c08, because the numbers in c01 are mostly kanji numerals and fall
outside the metric (Limitation 4).

\hypertarget{relation-to-prior-work}{%
\subsection{Relation to prior work}\label{relation-to-prior-work}}

LLM post-processing has been reported to degrade accurate extractions
\hypertarget{cite-gomezrey2026comparative-3}{}\protect\hyperlink{ref-gomezrey2026comparative}{{[}27{]}}. The
deterministic post-processing studied here did not worsen body CER or
core recall, in contrast to that failure mode; but LLM post-processing
was not measured, so we claim no superiority. The study also exposed the
limits of deterministic rules: where contextual judgment is needed, as
in heading hierarchy, rule-based reconstruction can overwrite the
converter's output and make it worse.

\hypertarget{why-human-checking-remains}{%
\subsection{Why human checking
remains}\label{why-human-checking-remains}}

A natural objection is that, when conversion fails, our answer reduces
to having someone look at the page, so that detection falls short of the
automation one would want. We respond with two points. First, what is
automated here is detection, but in this study it narrows the human's
task very little. With the notation removed, the token-level
disagreement signal marked 13.2\% of the number errors (Section 6.5;
54.6\% when digits inside notation were counted), with no token-level
false alarms. The review list that the pipeline currently writes (416
rows) aggregates by line, and none of the 24 number-only flags judged on
the sample was a real problem (Section 6.3). Most number errors are ones
that both engines share, which no disagreement can flag. Second,
automatic correction of the errors that matter most is blocked in
principle, not merely unimplemented. A digit converted into another
plausible digit, as in 1461 read as 1462, cannot be identified from the
output text alone, because the evidence for the true value exists only
in the source image. Assigning the checking role to a stronger engine
would only move the problem: that engine's own errors are unverifiable
in the same way, and an error that both engines make produces no
disagreement at all. Checking flagged items against the source image is
therefore not a stopgap for missing automation; it is the only step at
which a value is compared with the ground truth. What this study
measures is how little that step can currently be narrowed (about 13\%
of number errors are caught before it) and why (about 87\% are errors
that both engines make). This is the information a user needs in
deciding whether to keep critical figures under full source checking;
with this engine pair, they should.

\hypertarget{limitations}{%
\section{Limitations}\label{limitations}}

\begin{enumerate}
\def\labelenumi{\arabic{enumi}.}
\tightlist
\item
  \textbf{Small scale.} Three documents and 48 pages constitute only a
  preliminary evaluation. Document-level intervals are shown, but with
  only three documents they are wide and discrete, and they do not
  adequately capture between-document variation. c06 supplies 35 of the
  48 pages, so page means mostly reflect that one document (per-document
  means are in Section 6.1). Conclusions are limited to similar Japanese
  technical documents
  \hypertarget{cite-bergkirkpatrick2012significance-3}{}\protect\hyperlink{ref-bergkirkpatrick2012significance}{{[}33{]}},
  \hypertarget{cite-card2020power-3}{}\protect\hyperlink{ref-card2020power}{{[}34{]}}.
\item
  \textbf{Rasterized documents are not real scans.} They lack stains,
  skew, shadows, and fading, so the image condition represents a best
  case. Results on real scans may be worse.
\item
  \textbf{How the ground truth was made.} For all 48 pages, a language
  model corrected a text-layer draft against the page images; a person
  did not transcribe from the images alone. In the text-layer condition,
  the ground truth may inherit the errors and notations of the text
  layer and thus favor systems that output it. The 10\% blind
  re-transcription was carried out by a separate LLM assistant (Section
  4.6). Because the ground truth itself was made with LLM assistance,
  this check shows only agreement between two models; no independent
  check by a person has been done.
\item
  \textbf{The number-and-unit metric covers a narrow range.} Numbers in
  c01 are almost all kanji numerals and are not evaluated by a metric
  that counts only Arabic numerals (they are evaluated through body
  CER). For units outside the list, only the digits are counted. The
  number-and-unit results are in effect those for c06 and c08, and on
  some pages tick labels account for most of the tokens (reported
  separately as ``extra'' recall).
\item
  \textbf{The simplified-character replacement table has not been
  reviewed by a native Japanese speaker.} Detection counts are detector
  outputs, not ground truth. Simplified forms that lie within JIS X 0208
  but are distinct characters in Japanese (about 250 characters) cannot
  be detected at the character level.
\item
  \textbf{Two engines that make the same error defeat the check.} We
  report the miss rate, but its value depends on the documents and the
  engine pair. The measured miss rate is large (85 to 87\% with the
  notation removed), and most misses are numbers that neither engine
  reads.
\item
  \textbf{E exists only in the rasterized condition.} The production
  pipeline requires OCR-path input, so no post-processed output exists
  for the text-layer condition, and the effect of post-processing is
  measured on rasterized documents only. Side effects must be checked
  visually. The accuracy of formula recognition itself is out of scope.
\item
  \textbf{System B is a text-only OCR engine.} NDLOCR-Lite returns plain
  text without Markdown structure, so its heading and table scores are
  structurally lower than those of a converter that outputs structure.
  It is included as an OCR baseline, and its results should be read
  accordingly.
\item
  \textbf{MinerU configuration.} Because Japanese was removed from the
  OCR language choices in version 3.4, the configuration may
  disadvantage Japanese. A comparison with the 3.3 series (the last
  version in which Japanese can be chosen) was not run in this study; it
  is listed in the extension plan in Section 9.
\item
  \textbf{Visually classified failures (F2, F5).} The visual
  classification was not carried out in this run (item 13(c)), so F2 and
  F5 of the taxonomy remain untested here. The design has a single
  annotator; agreement can be reported only after a second
  classification.
\item
  \textbf{Out of scope.} Vertical writing, ruby, and old character forms
  are excluded. The baselines are MinerU and NDLOCR-Lite; Docling,
  Marker, YomiToku and others are not included.
\item
  \textbf{The literature survey} is based on search. Unverified
  references are not cited. Statements that we ``found no'' prior work
  are not based on an exhaustive survey.
\item
  \textbf{Observed during execution (added 2026-11-08).} (a) The
  script-contamination metric fails silently when the scoring program
  cannot execute the guard CLI, and it was therefore absent from the
  first scoring run; the values in this paper were recomputed with the
  same CLI and cross-checked in two environments. (b) E's input was a
  copy of A's MinerU output with unchanged content, in which the method
  folder was renamed from \texttt{auto} to \texttt{ocr} to match the
  form the production runner accepts; the identity of the content is
  preserved in a form that can be verified by hash. (c) The visual
  classification of F2 and F5 was not carried out in this run, which is
  why Section 6.6 reports them as not judged rather than not reproduced.
  (d) Processing time was not recorded, so this paper makes no speed
  comparison.
\item
  \textbf{Post-hoc scorer change (added 2026-11-10).} The pre-specified
  number counter reads digits from the whole output, including notation
  in the converter's Markdown (image-link paths, HTML attributes,
  comments). This inflated A's token count and made the pre-specified F1
  favor B. A notation-free variant (\texttt{metrics.strip\_markup}) is
  reported next to it (Section 6.1) and shows no difference between A
  and B. The variant was chosen after the results were seen, and the set
  of what counts as notation (HTML tags and comments, image-link paths,
  link targets) is our choice. The disagreement check (H2) also counted
  digits in the raw output; it was rerun with \texttt{-\/-strip-markup},
  and its detection rate fell from 54.6\% to 13.2\% (Section 6.5).
\end{enumerate}

\hypertarget{conclusion}{%
\section{Conclusion}\label{conclusion}}

We converted three Japanese technical documents (48 pages) under two
conditions (text layer and rasterized), and scored numbers with units
exhaustively. NDLOCR-Lite (B) had lower body CER than MinerU (A) in the
page means (0.012 against 0.053 rasterized) and returned no structure.
Its higher pre-specified number-and-unit F1 (0.575 against 0.397
rasterized) came from digits in the notation of A's Markdown (image-link
paths, HTML attributes); with the notation removed, a post-hoc check,
the two tie (0.575 against 0.594). The production post-processing (E)
raised the recall of extra numbers by +0.145 without degrading body
text, but this gain consists mostly of page numbers and footnote text
that the pipeline restores, not of figure values. Heading-level
agreement fell from 0.133 to 0.061 (the fall lies entirely in c08) and
spurious formulas stayed at 3; partner-gated replacement never fired, so
the partner and no-partner outputs were identical. Disagreement
detection caught 13.2\% of number errors (rasterized; 54.6\% when digits
inside notation were counted) with no false alarms at token level.
Verdicts: H1 was not supported (the flagging-precision gate passed at
0.929, but the heading and spurious-formula items missed their
criteria); H2 had no threshold set in advance, and its detection rate
(13.2\%) was far short of ``most''; and H3 was partly supported (F1 and
F4 reproduced, F3 below criterion, F2 and F5 not judged). Under the
criteria fixed in advance, the outcome is to stay a technical report. We
recommend scoring numbers only after removing notation that is not page
content, and checking critical figures against the source.

We plan to extend the study to 100--300 pages; to add Docling, Marker,
and NDLOCR baselines; to include real scans (the two CC BY papers from
volume 40 of the \emph{Journal of the Society of Agricultural
Structures, Japan}, c10 and c11 in \texttt{docs/candidates.md}; licenses
to be re-checked), the formula-rich c09, and the MinerU 3.3 series; and
to have a native speaker review the replacement table.

Behind these extensions lies a larger task: a benchmark specialized to
Japanese technical documents, assembled under licenses that permit
redistribution. This study is a first step toward it, and our own
scoring pitfall, counting digits in the converter's notation as number
errors, shows why such a benchmark must define what a number token is.

The converters' raw outputs are not released, and the text reports
aggregate scores and short excerpts. The manuscript is under CC BY-NC-ND
4.0.

\hypertarget{appendices}{%
\section{Appendices}\label{appendices}}

\hypertarget{appendix-a-post-processing-pipeline-and-partner-settings}{%
\subsection{Appendix A: Post-processing pipeline and partner
settings}\label{appendix-a-post-processing-pipeline-and-partner-settings}}

Configuration E is the PDFtoMD pipeline (\texttt{pipeline/run.py}), run
per page on A's MinerU output with \texttt{-\/-partner} set to the
page's NDLOCR-Lite text and \texttt{-\/-text-only-partner} to the page's
Tesseract text (language \texttt{jpn}, page segmentation mode 4). The
ablation omits both partner arguments. Per-page review lists
(\texttt{review.csv}) are generated for every page; they are not
published, because they quote the source documents at length (Section
9), and their aggregate counts are reported in Section 6. Pipeline
version: \texttt{PDFtoMD\_HighAccuracy\_proto\_v0.1.0} (production
prototype).

\hypertarget{appendix-b-transcription-rules}{%
\subsection{Appendix B: Transcription
rules}\label{appendix-b-transcription-rules}}

The transcription rules are given in
\texttt{docs/transcription-rules.md} in the project repository, in the
version fixed when the ground truth was made. They state which
characters are not distinguished after NFKC normalization and whitespace
removal, which are distinguished, and the rules for numbers and units,
tables, and headings.

\hypertarget{appendix-c-post-processing-components}{%
\subsection{Appendix C: Post-processing
components}\label{appendix-c-post-processing-components}}

\begin{longtable}[]{@{}
  >{\raggedright\arraybackslash}p{(\columnwidth - 4\tabcolsep) * \real{0.3333}}
  >{\raggedright\arraybackslash}p{(\columnwidth - 4\tabcolsep) * \real{0.3333}}
  >{\raggedright\arraybackslash}p{(\columnwidth - 4\tabcolsep) * \real{0.3333}}@{}}
\caption{Post-processing components.}\tabularnewline
\toprule\noalign{}
\begin{minipage}[b]{\linewidth}\raggedright
Component
\end{minipage} & \begin{minipage}[b]{\linewidth}\raggedright
Content
\end{minipage} & \begin{minipage}[b]{\linewidth}\raggedright
Failure
\end{minipage} \\
\midrule\noalign{}
\endfirsthead
\toprule\noalign{}
\begin{minipage}[b]{\linewidth}\raggedright
Component
\end{minipage} & \begin{minipage}[b]{\linewidth}\raggedright
Content
\end{minipage} & \begin{minipage}[b]{\linewidth}\raggedright
Failure
\end{minipage} \\
\midrule\noalign{}
\endhead
\bottomrule\noalign{}
\endlastfoot
P1/P2 anchors, footnotes & page-number and footnote items from the
content list placed back into the Markdown & page correspondence \\
P6 heading hierarchy & levels rebuilt from the content list & F4 \\
P3 formula repair & replacement only when both partners agree against
the main engine; otherwise flagged & F3 \\
P4 character fixes & small logged set of unambiguous corrections & F2 \\
Flagging & suspected simplified/non-JIS characters, number
disagreements, unconfirmed formula spans written to \texttt{review.csv};
simplified characters are flagged, never replaced (table unreviewed) &
F1, F3 \\
\end{longtable}

\hypertarget{appendix-d-evaluation-documents-source-retrieval-date-sha-256}{%
\subsection{Appendix D: Evaluation documents (source, retrieval date,
SHA-256)}\label{appendix-d-evaluation-documents-source-retrieval-date-sha-256}}

\textbf{c01} --- 4 pages. Retrieved 2026-10-29. URL:
\url{https://www.mlit.go.jp/notice/noticedata/pdf/201703/00006505.pdf}
SHA-256:
\texttt{2ac4331fd5bfb4a29fbd9fa8610867db8a97665a2c3d34c1b6f6a561d2068552}

\textbf{c06} --- 35 pages. Retrieved 2026-10-29. URL:
\url{https://www.jstage.jst.go.jp/static/files/en/pub_survey2024_society.pdf}
SHA-256:
\texttt{f6078bfeae0e96969552acf7edd6e63f6a0e8c0e07212d0f43a1b3669afb25c4}

\textbf{c08} --- 9 pages. Retrieved 2026-10-30. URL:
\url{https://www.jstage.jst.go.jp/article/crt/37/0/37_47/_pdf/-char/ja}
SHA-256:
\texttt{6f6dcb0d923d4a9ba15b1c4930cf043393258ecded213837cff2ec47d7dceb25}

Rasterized versions (c01i, c06i, c08i) are these PDFs converted to
image-only PDFs at 300 dpi with \texttt{tools/prep/rasterize.py}; their
hashes and the license evidence (column \texttt{license\_evidence}) are
in \texttt{data/manifest.csv}.

\hypertarget{appendix-e-execution-environment}{%
\subsection{Appendix E: Execution
environment}\label{appendix-e-execution-environment}}

\begin{longtable}[]{@{}
  >{\raggedright\arraybackslash}p{(\columnwidth - 2\tabcolsep) * \real{0.5000}}
  >{\raggedright\arraybackslash}p{(\columnwidth - 2\tabcolsep) * \real{0.5000}}@{}}
\caption{Execution environment.}\tabularnewline
\toprule\noalign{}
\begin{minipage}[b]{\linewidth}\raggedright
Item
\end{minipage} & \begin{minipage}[b]{\linewidth}\raggedright
Content
\end{minipage} \\
\midrule\noalign{}
\endfirsthead
\toprule\noalign{}
\begin{minipage}[b]{\linewidth}\raggedright
Item
\end{minipage} & \begin{minipage}[b]{\linewidth}\raggedright
Content
\end{minipage} \\
\midrule\noalign{}
\endhead
\bottomrule\noalign{}
\endlastfoot
MinerU & 3.4.5, pipeline backend, default settings (conversion method
not fixed), run date 2026-11-08 \\
NDLOCR-Lite & 1.3.0 (git \texttt{e57f7c6}), run date 2026-11-08 \\
Tesseract & 5.5.3, language \texttt{jpn}, psm 4 (partner for E; not
scored as a system) \\
Post-processing (E) & PDFtoMD pipeline, version
\texttt{PDFtoMD\_HighAccuracy\_proto\_v0.1.0}, run date 2026-11-08 \\
OS, GPU & macOS 26.6.2 on an Apple M1 MacBook Air (arm64) \\
Scoring code commit & the tools of this repository as of the commit that
records the results (\texttt{fa247948}) \\
Scoring command &
\texttt{python3\ tools/ocr-eval/run\_eval.py\ -\/-systems\ ...\ -\/-baseline\ mineru} \\
\end{longtable}

\hypertarget{references}{%
\section*{References}\label{references}}
\addcontentsline{toc}{section}{References}

\hypertarget{refs}{}
\begin{CSLReferences}{0}{0}
\leavevmode\vadjust pre{\hypertarget{ref-wang2024mineru}{}}%
\CSLLeftMargin{{[}1{]} }%
\CSLRightInline{B. Wang, C. Xu, \emph{et al.}, {``MinerU: An open-source
solution for precise document content extraction,''} \emph{arXiv
preprint}, 2024, Available: \url{https://arxiv.org/abs/2409.18839}}

\leavevmode\vadjust pre{\hypertarget{ref-auer2024docling}{}}%
\CSLLeftMargin{{[}2{]} }%
\CSLRightInline{C. Auer \emph{et al.}, {``Docling technical report,''}
\emph{arXiv preprint}, 2024, Available:
\url{https://arxiv.org/abs/2408.09869}}

\leavevmode\vadjust pre{\hypertarget{ref-poznanski2025olmocr}{}}%
\CSLLeftMargin{{[}3{]} }%
\CSLRightInline{J. Poznanski \emph{et al.}, {``olmOCR: Unlocking
trillions of tokens in PDFs with vision language models,''} \emph{arXiv
preprint}, 2025, Available: \url{https://arxiv.org/abs/2502.18443}}

\leavevmode\vadjust pre{\hypertarget{ref-ouyang2025omnidocbench}{}}%
\CSLLeftMargin{{[}4{]} }%
\CSLRightInline{L. Ouyang \emph{et al.}, {``OmniDocBench: Benchmarking
diverse PDF document parsing with comprehensive annotations,''} in
\emph{Proceedings of the IEEE/CVF conference on computer vision and
pattern recognition (CVPR)}, 2025. Available:
\url{https://arxiv.org/abs/2412.07626}}

\leavevmode\vadjust pre{\hypertarget{ref-stockmark2026jasyn}{}}%
\CSLLeftMargin{{[}5{]} }%
\CSLRightInline{Stockmark Inc., {``OmniDocBench-JASyn: A synthetic
{Japanese} document parsing benchmark in OmniDocBench format.''} Hugging
Face dataset, 2026. Available:
\url{https://huggingface.co/datasets/stockmark/OmniDocBench-JASyn}}

\leavevmode\vadjust pre{\hypertarget{ref-sasagawa2026synthjdoc}{}}%
\CSLLeftMargin{{[}6{]} }%
\CSLRightInline{K. Sasagawa \emph{et al.}, {``Synth-JDoc: Synthesizing a
{Japanese} document image dataset for OCR with diverse layouts and
embedded images,''} \emph{arXiv preprint}, 2026, Available:
\url{https://arxiv.org/abs/2608.28248}}

\leavevmode\vadjust pre{\hypertarget{ref-sasagawa2026vertical}{}}%
\CSLLeftMargin{{[}7{]} }%
\CSLRightInline{K. Sasagawa, S. Kurita, and D. Kawahara, {``Evaluating
multimodal large language models on vertically written {Japanese}
text,''} in \emph{Proceedings of LREC 2026}, 2026. Available:
\url{https://aclanthology.org/2026.lrec-main.713/}}

\leavevmode\vadjust pre{\hypertarget{ref-zhong2020pubtabnet}{}}%
\CSLLeftMargin{{[}8{]} }%
\CSLRightInline{X. Zhong, E. ShafieiBavani, and A. Jimeno Yepes,
{``Image-based table recognition: Data, model, and evaluation,''} in
\emph{Computer vision -- ECCV 2020}, 2020. doi:
\href{https://doi.org/10.1007/978-3-030-58589-1_34}{10.1007/978-3-030-58589-1\_34}.}

\leavevmode\vadjust pre{\hypertarget{ref-wang2025cdm}{}}%
\CSLLeftMargin{{[}9{]} }%
\CSLRightInline{B. Wang \emph{et al.}, {``Image over text: Transforming
formula recognition evaluation with character detection matching,''} in
\emph{Proceedings of the IEEE/CVF conference on computer vision and
pattern recognition (CVPR)}, 2025. Available:
\url{https://arxiv.org/abs/2409.03643}}

\leavevmode\vadjust pre{\hypertarget{ref-li2024readoc}{}}%
\CSLLeftMargin{{[}10{]} }%
\CSLRightInline{Z. Li \emph{et al.}, {``READoc: A unified benchmark for
realistic document structured extraction,''} \emph{arXiv preprint},
2024, Available: \url{https://arxiv.org/abs/2409.05137}}

\leavevmode\vadjust pre{\hypertarget{ref-he2025fincriticaled}{}}%
\CSLLeftMargin{{[}11{]} }%
\CSLRightInline{Y. He, X. Peng, \emph{et al.}, {``FinCriticalED: A
visual benchmark for financial fact-level OCR evaluation,''} \emph{arXiv
preprint}, 2025, Available: \url{https://arxiv.org/abs/2511.14998}}

\leavevmode\vadjust pre{\hypertarget{ref-zhang2024ohrbench}{}}%
\CSLLeftMargin{{[}12{]} }%
\CSLRightInline{J. Zhang, Q. Zhang, B. Wang, \emph{et al.}, {``{OCR}
hinders {RAG}: Evaluating the cascading impact of {OCR} on
retrieval-augmented generation,''} \emph{arXiv preprint}, 2024,
Available: \url{https://arxiv.org/abs/2412.02592}}

\leavevmode\vadjust pre{\hypertarget{ref-livathinos2025docling}{}}%
\CSLLeftMargin{{[}13{]} }%
\CSLRightInline{N. Livathinos \emph{et al.}, {``Docling: An efficient
open-source toolkit for AI-driven document conversion,''} \emph{arXiv
preprint}, 2025, Available: \url{https://arxiv.org/abs/2501.17887}}

\leavevmode\vadjust pre{\hypertarget{ref-yomitoku2024}{}}%
\CSLLeftMargin{{[}14{]} }%
\CSLRightInline{K. Kinoshita, {``YomiToku: {Japanese} document AI engine
(OCR and layout analysis).''}
\url{https://github.com/kotaro-kinoshita/yomitoku}, 2024.}

\leavevmode\vadjust pre{\hypertarget{ref-ndlocr_cli}{}}%
\CSLLeftMargin{{[}15{]} }%
\CSLRightInline{National Diet Library, Japan (NDL Lab), {``NDLOCR
(ver.2.1) application repository.''}
\url{https://github.com/ndl-lab/ndlocr_cli}, 2023.}

\leavevmode\vadjust pre{\hypertarget{ref-ndlocr_lite}{}}%
\CSLLeftMargin{{[}16{]} }%
\CSLRightInline{National Diet Library, Japan (NDL Lab),
{``NDLOCR-lite.''} \url{https://github.com/ndl-lab/ndlocr-lite}, 2026.}

\leavevmode\vadjust pre{\hypertarget{ref-mineru2026software}{}}%
\CSLLeftMargin{{[}17{]} }%
\CSLRightInline{OpenDataLab, {``MinerU (version 3.4.5).''}
\url{https://github.com/opendatalab/MinerU}, 2026.}

\leavevmode\vadjust pre{\hypertarget{ref-onami2024jdocqa}{}}%
\CSLLeftMargin{{[}18{]} }%
\CSLRightInline{E. Onami, S. Kurita, T. Miyanishi, and T. Watanabe,
{``{JDocQA}: {Japanese} document question answering dataset for
generative language models,''} in \emph{Proceedings of LREC-COLING
2024}, 2024. Available: \url{https://arxiv.org/abs/2403.19454}}

\leavevmode\vadjust pre{\hypertarget{ref-nagata1998japanese}{}}%
\CSLLeftMargin{{[}19{]} }%
\CSLRightInline{M. Nagata, {``Japanese {OCR} error correction using
character shape similarity and statistical language model,''} in
\emph{Proc. COLING-ACL '98}, Montr{é}al, 1998, pp. 922--928. doi:
\href{https://doi.org/10.3115/980691.980721}{10.3115/980691.980721}.}

\leavevmode\vadjust pre{\hypertarget{ref-masuda2015global}{}}%
\CSLLeftMargin{{[}20{]} }%
\CSLRightInline{K. Masuda, {``{OCR} character error correction using
global information (in {Japanese}),''} in \emph{Proceedings of the 21st
annual meeting of the association for natural language processing
(NLP2015)}, 2015, pp. P2--2. Available:
\url{https://anlp.jp/proceedings/annual_meeting/2015/pdf_dir/P2-2.pdf}}

\leavevmode\vadjust pre{\hypertarget{ref-paddleocr_issue11249}{}}%
\CSLLeftMargin{{[}21{]} }%
\CSLRightInline{madmalkav and matt-m-o, {``25 jōyō kanji (regular-use
kanji) not included in japanese model.''} PaddlePaddle/PaddleOCR issue
\#11249 / discussion \#12940,
\url{https://github.com/PaddlePaddle/PaddleOCR/issues/11249}, 2023.}

\leavevmode\vadjust pre{\hypertarget{ref-mineru_issue2521}{}}%
\CSLLeftMargin{{[}22{]} }%
\CSLRightInline{lovetian1991, {``Japanese language support is not
working well (issue title in {Chinese}).''} opendatalab/MinerU issue
\#2521, \url{https://github.com/opendatalab/MinerU/issues/2521}, 2025.}

\leavevmode\vadjust pre{\hypertarget{ref-karita2023lenient}{}}%
\CSLLeftMargin{{[}23{]} }%
\CSLRightInline{S. Karita, R. Sproat, and H. Ishikawa, {``Lenient
evaluation of {Japanese} speech recognition: Modeling naturally
occurring spelling inconsistency,''} in \emph{Proceedings of the ACL
workshop on computation and written language (CAWL)}, 2023. Available:
\url{https://arxiv.org/abs/2306.04530}}

\leavevmode\vadjust pre{\hypertarget{ref-nguyen2021survey}{}}%
\CSLLeftMargin{{[}24{]} }%
\CSLRightInline{T. T. H. Nguyen, A. Jatowt, M. Coustaty, and A. Doucet,
{``Survey of post-{OCR} processing approaches,''} \emph{ACM Computing
Surveys}, vol. 54, no. 6, pp. 124:1--124:37, 2021, doi:
\href{https://doi.org/10.1145/3453476}{10.1145/3453476}.}

\leavevmode\vadjust pre{\hypertarget{ref-chiron2017icdar}{}}%
\CSLLeftMargin{{[}25{]} }%
\CSLRightInline{G. Chiron, A. Doucet, M. Coustaty, and J.-P. Moreux,
{``{ICDAR2017} competition on post-{OCR} text correction,''} in
\emph{Proc. 14th IAPR int. Conf. On document analysis and recognition
(ICDAR)}, IEEE, 2017, pp. 1423--1428. doi:
\href{https://doi.org/10.1109/ICDAR.2017.232}{10.1109/ICDAR.2017.232}.}

\leavevmode\vadjust pre{\hypertarget{ref-kanerva2025nofreelunches}{}}%
\CSLLeftMargin{{[}26{]} }%
\CSLRightInline{J. Kanerva, C. Ledins, S. Käpyaho, and F. Ginter,
{``{OCR} error post-correction with {LLMs} in historical documents: No
free lunches,''} in \emph{Proc. Third workshop on resources and
representations for under-resourced languages and domains
(RESOURCEFUL-2025)}, 2025. Available:
\url{https://arxiv.org/abs/2502.01205}}

\leavevmode\vadjust pre{\hypertarget{ref-gomezrey2026comparative}{}}%
\CSLLeftMargin{{[}27{]} }%
\CSLRightInline{M. Gómez Rey, P. Callejo, M. Muñoz-Organero, and C.
Alario-Hoyos, {``A comparative evaluation of digitization pipelines for
historiographical sources.''} 2026. Available:
\url{https://arxiv.org/abs/2608.24976}}

\leavevmode\vadjust pre{\hypertarget{ref-fiscus1997rover}{}}%
\CSLLeftMargin{{[}28{]} }%
\CSLRightInline{J. G. Fiscus, {``A post-processing system to yield
reduced word error rates: Recognizer output voting error reduction
({ROVER}),''} in \emph{Proc. IEEE workshop on automatic speech
recognition and understanding (ASRU)}, 1997, pp. 347--354. doi:
\href{https://doi.org/10.1109/ASRU.1997.659110}{10.1109/ASRU.1997.659110}.}

\leavevmode\vadjust pre{\hypertarget{ref-lund2009improving}{}}%
\CSLLeftMargin{{[}29{]} }%
\CSLRightInline{W. B. Lund and E. K. Ringger, {``Improving optical
character recognition through efficient multiple system alignment,''} in
\emph{Proc. 9th ACM/IEEE-CS joint conference on digital libraries
(JCDL)}, 2009, pp. 231--240. doi:
\href{https://doi.org/10.1145/1555400.1555437}{10.1145/1555400.1555437}.}

\leavevmode\vadjust pre{\hypertarget{ref-lund2014generalize}{}}%
\CSLLeftMargin{{[}30{]} }%
\CSLRightInline{W. B. Lund, E. K. Ringger, and D. D. Walker, {``How well
does multiple {OCR} error correction generalize?''} in \emph{Document
recognition and retrieval XXI (DRR 2014), proc. SPIE 9021}, 2014, p.
90210A. doi:
\href{https://doi.org/10.1117/12.2042502}{10.1117/12.2042502}.}

\leavevmode\vadjust pre{\hypertarget{ref-reul2018pretraining}{}}%
\CSLLeftMargin{{[}31{]} }%
\CSLRightInline{C. Reul, U. Springmann, C. Wick, and F. Puppe,
{``Improving {OCR} accuracy on early printed books by combining
pretraining, voting, and active learning,''} \emph{Journal for Language
Technology and Computational Linguistics (JLCL)}, vol. 33, no. 1, 2018,
Available: \url{https://arxiv.org/abs/1802.10038}}

\leavevmode\vadjust pre{\hypertarget{ref-koehn2004significance}{}}%
\CSLLeftMargin{{[}32{]} }%
\CSLRightInline{P. Koehn, {``Statistical significance tests for machine
translation evaluation,''} in \emph{Proceedings of EMNLP 2004},
Barcelona, Spain, 2004, pp. 388--395. Available:
\url{https://aclanthology.org/W04-3250/}}

\leavevmode\vadjust pre{\hypertarget{ref-bergkirkpatrick2012significance}{}}%
\CSLLeftMargin{{[}33{]} }%
\CSLRightInline{T. Berg-Kirkpatrick, D. Burkett, and D. Klein, {``An
empirical investigation of statistical significance in {NLP},''} in
\emph{Proceedings of EMNLP-CoNLL 2012}, 2012, pp. 995--1005. Available:
\url{https://aclanthology.org/D12-1091/}}

\leavevmode\vadjust pre{\hypertarget{ref-card2020power}{}}%
\CSLLeftMargin{{[}34{]} }%
\CSLRightInline{D. Card, P. Henderson, U. Khandelwal, R. Jia, K.
Mahowald, and D. Jurafsky, {``With little power comes great
responsibility,''} in \emph{Proceedings of EMNLP 2020}, 2020, pp.
9263--9274. doi:
\href{https://doi.org/10.18653/v1/2020.emnlp-main.745}{10.18653/v1/2020.emnlp-main.745}.}

\leavevmode\vadjust pre{\hypertarget{ref-vanmiltenburg2021preregistering}{}}%
\CSLLeftMargin{{[}35{]} }%
\CSLRightInline{E. van Miltenburg, C. van der Lee, and E. Krahmer,
{``Preregistering {NLP} research,''} in \emph{Proceedings of NAACL-HLT
2021}, 2021, pp. 613--623. doi:
\href{https://doi.org/10.18653/v1/2021.naacl-main.51}{10.18653/v1/2021.naacl-main.51}.}

\leavevmode\vadjust pre{\hypertarget{ref-shi2017rctw}{}}%
\CSLLeftMargin{{[}36{]} }%
\CSLRightInline{B. Shi \emph{et al.}, {``ICDAR2017 competition on
reading {Chinese} text in the wild (RCTW-17),''} \emph{arXiv preprint},
2017, Available: \url{https://arxiv.org/abs/1708.09585}}

\leavevmode\vadjust pre{\hypertarget{ref-wemhoener2013improved}{}}%
\CSLLeftMargin{{[}37{]} }%
\CSLRightInline{D. Wemhoener, I. Z. Yalniz, and R. Manmatha, {``Creating
an improved version using noisy {OCR} from multiple editions,''} in
\emph{Proc. 12th int. Conf. On document analysis and recognition
(ICDAR)}, 2013, pp. 160--164. doi:
\href{https://doi.org/10.1109/ICDAR.2013.39}{10.1109/ICDAR.2013.39}.}

\end{CSLReferences}

\end{document}